\documentclass[sigconf,nonacm]{acmart}

\usepackage{pifont}
\usepackage{enumitem}
\usepackage[most]{tcolorbox}
\usepackage{multirow}
\usepackage{booktabs}      
\usepackage{threeparttable} 

\usepackage{algorithm}
\usepackage{algpseudocode}

\usepackage{tcolorbox}
\tcbuselibrary{skins,breakable}
\usepackage{chngcntr}

\usepackage{tikz}
\usetikzlibrary{
    arrows.meta,
    positioning,
    calc,
    fit,
    shapes,
    shapes.geometric
}

\usepackage{booktabs}
\usepackage[table]{xcolor}

\newlist{promptitemize}{itemize}{2}

\setlist[promptitemize,1]{
    label=\textbullet,
    leftmargin=1.5em,
    itemsep=0.15em,
    topsep=0.25em,
    parsep=0pt,
    partopsep=0pt
}

\setlist[promptitemize,2]{
    label=--,
    leftmargin=1.5em,
    itemsep=0.15em,
    topsep=0.15em,
    parsep=0pt,
    partopsep=0pt
}

\newlist{promptenum}{enumerate}{2}

\setlist[promptenum,1]{
    label=\arabic*.,
    leftmargin=1.8em,
    itemsep=0.25em,
    topsep=0.25em,
    parsep=0pt,
    partopsep=0pt
}

\setlist[promptenum,2]{
    label=\alph*),
    leftmargin=1.8em,
    itemsep=0.15em,
    topsep=0.15em,
    parsep=0pt,
    partopsep=0pt
}

\newtcolorbox{promptbox}[1]{
    enhanced jigsaw,
    breakable,
    colback=white,
    colframe=gray!65,
    colbacktitle=gray!65,
    coltitle=white,
    fonttitle=\bfseries,
    fontupper=\small,
    title={#1},
    title after break={#1\ (continued)},
    boxrule=0.8pt,
    arc=1.5mm,
    outer arc=1.5mm,
    left=6mm,
    right=6mm,
    top=3mm,
    bottom=3mm,
    toptitle=0.2mm,
    bottomtitle=0.2mm,
    lefttitle=3mm,
    righttitle=3mm,
    before skip=7pt,
    after skip=7pt
}

\AtBeginDocument{%
  }

\begin{document}

\settopmatter{printacmref=false}
\setcopyright{none}
\renewcommand\footnotetextcopyrightpermission[1]{}
\pagestyle{plain}

\title[FedEHR-Agents]{FedEHR-Agents: Federated Agentic Optimization for Automated EHR Modeling}

\makeatletter
\let\OriginalClassWarning\ClassWarning
\renewcommand{\ClassWarning}[2]{}
\pdfstringdefDisableCommands{%
  \def\textsuperscript#1{#1}%
  \def\textbf#1{#1}%
}

\author{%
\textbf{
Jun Bai\textsuperscript{1,2},
Ruilin Wang\textsuperscript{1,2},
Yue Li\textsuperscript{1,2,*}
}
}

\let\ClassWarning\OriginalClassWarning
\makeatother

\affiliation{%
  \institution{%
    \textsuperscript{1}School of Computer Science, McGill University,
    Montreal, Canada\\
    \textsuperscript{2}Mila -- Quebec AI Institute,
    Montreal, Canada\\
  }
  \city{\unskip}
  \country{\unskip}
}

\begin{abstract}
Recent advances in large language models are enabling autonomous clinical agents to perform increasingly complex electronic health record (EHR) modeling workflows. However, agents deployed at individual hospitals remain constrained by institution-specific data and modeling environments, while direct cross-hospital collaboration is restricted by the sensitivity of patient-level EHR data. Although federated learning (FL) provides a natural foundation for privacy-preserving collaboration, existing approaches remain predominantly model-centric, limiting federation to prediction models or their updates while overlooking the richer modeling experience accumulated by autonomous agents. To address this limitation, we propose \textbf{FedEHR-Agents}, an experience-centric federated agentic optimization framework for automated EHR modeling. Each hospital deploys an autonomous clinical EHR agent that performs data preprocessing and model development while refining local clinical modeling experience through historical memory, task-specific evaluation, and TextGrad-based prompt refinement. The federated server performs evidence-guided experience aggregation to integrate reliable and complementary modeling experience across heterogeneous hospitals and distills the aggregated experience into global meta-prompts for subsequent local refinement. Extensive experiments on real-world multi-hospital EHR benchmarks demonstrate that FedEHR-Agents consistently outperforms local and federated baselines across diverse clinical prediction tasks and remains robust across different federation scales and LLM backbones. These results establish clinical modeling experience as a promising collaborative object beyond conventional parameter-centric FL and point toward federated autonomous clinical intelligence.
\end{abstract}


\maketitle
\begingroup
\renewcommand{\thefootnote}{*}
\begin{NoHyper}
\footnotetext{Corresponding author: Yue Li (yueli@cs.mcgill.ca)}
\end{NoHyper}
\endgroup

\section{Introduction} \label{sec:introduction}

\begin{figure}[!t]
\centering
\setlength{\abovecaptionskip}{0.3cm}
\includegraphics[width=0.95\linewidth,scale=1.0]{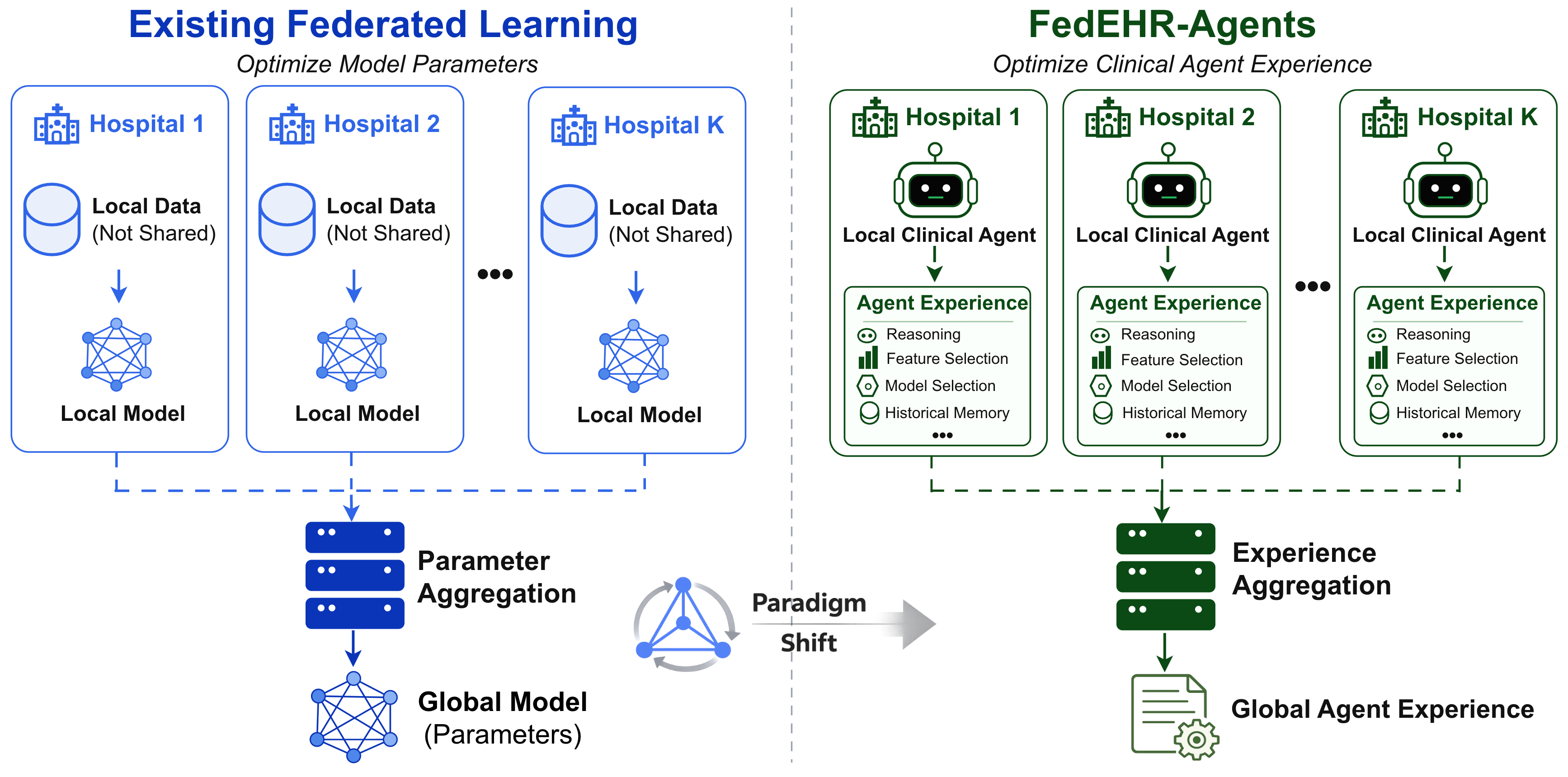}
\caption{
Motivation of FedEHR-Agents. Conventional FL performs collaboration through prediction model parameter aggregation, whereas FedEHR-Agents shifts the collaborative object toward clinical modeling experience.
}
\Description{.}
\label{fig:motivation}
\vspace{-0.4cm}
\end{figure}

Electronic health records (EHRs) have become a major data source for developing clinical prediction systems, supporting applications such as mortality prediction, sepsis detection, length-of-stay estimation, and disease progression modeling \cite{johnson2016mimic,pollard2018eicu,shickel2017deep,rajkomar2018scalable}. Meanwhile, recent advances in large language models (LLMs) are extending clinical AI beyond task-specific prediction toward more autonomous and multi-stage workflows \cite{teo2025generative,ren2025healthcare,ferber2026towards}. Equipped with reasoning, planning, memory, reflection, and tool-use capabilities \cite{yao2023react,shinn2023reflexion,guo2024llmagent,yuksekgonul2024textgrad}, LLM-based agents provide a promising foundation for automating complex modeling processes. Emerging biomedical agentic systems further demonstrate the feasibility of autonomously developing and iteratively improving complete machine learning solutions \cite{martinek2026agentomics,wang2025agent,wang2026doctoragents}. These advances suggest a transition from isolated prediction models toward autonomous clinical agents capable of managing broader EHR modeling workflows.

However, deploying such agents independently within individual hospitals inherently limits the modeling knowledge they can acquire. EHR data are highly heterogeneous across institutions because of differences in patient populations, clinical workflows, measurement practices, and documentation systems \cite{hunik2025diagnostic}. Consequently, agents operating in different hospitals encounter distinct data characteristics and modeling challenges, and may accumulate complementary preprocessing strategies, feature engineering decisions, and model development experience. Enabling these agents to collaboratively learn from such complementary experience could substantially improve the robustness and generalizability of automated clinical modeling across institutions. Direct collaboration, however, is constrained by the sensitive nature of clinical data: pooling patient-level EHRs across hospitals is restricted by privacy regulations and institutional governance requirements \cite{kaissis2020secure,rieke2020future}. This creates a fundamental need for privacy-preserving collaboration among autonomous clinical agents.

Federated learning (FL) provides a natural foundation for such collaboration by allowing hospitals to jointly develop machine learning models while keeping raw patient data local \cite{mcmahan2017fedavg,bai2025unified}. Considerable research has improved federated optimization under heterogeneous data distributions, including FedProx \cite{li2020fedprox}, SCAFFOLD \cite{karimireddy2020scaffold}, FedNova \cite{wang2020fednova}, and FedDyn \cite{acar2021feddyn}. Nevertheless, despite substantial advances in optimization robustness and convergence, the collaborative object of conventional FL remains primarily the prediction model or its parameter updates \cite{zhu2025fedweight}. Such a model-centric paradigm captures only the final outcome of clinical model development. The broader decisions that produce this outcome---including data preprocessing, feature engineering, model selection, performance evaluation, and iterative refinement---are typically made independently within each hospital and remain outside the federated optimization loop. Consequently, existing FL enables collaborative model learning but provides limited support for collaborative improvement of the agents that construct these models.

This limitation reveals a largely unexplored opportunity for federated clinical agents (Fig.~\ref{fig:motivation}). Developing a high-quality clinical prediction system requires a sequence of reasoning-driven decisions that substantially influence downstream performance \cite{shickel2017deep,holzinger2019causability,topol2019high}. Throughout this process, autonomous agents can accumulate valuable \emph{clinical modeling experience}, including effective preprocessing strategies, feature engineering decisions, model preferences, execution feedback, and historical refinement traces. Unlike prediction model parameters, such experience captures reusable knowledge about \emph{how} clinical models are developed and improved, and may therefore be transferable across heterogeneous hospitals without exchanging patient-level EHR data. Existing FL methods, however, neither explicitly preserve nor collaboratively optimize this accumulated experience. This raises a fundamental question: \emph{how can distributed clinical agents collaboratively accumulate, refine, and transfer clinical modeling experience across hospitals while preserving patient data locality?}

To address this challenge, we propose \textbf{FedEHR-Agents}, a federated agentic optimization framework for automated EHR modeling that shifts the collaborative objective from prediction model parameters to clinical modeling experience. Each hospital deploys an autonomous clinical EHR agent to perform data preprocessing and model development over EHR data, while continuously accumulating and refining local modeling experience through historical memory, task-specific evaluation, and TextGrad-based prompt refinement (Fig.~\ref{fig:fedehr-agents}). Rather than exchanging prediction models, each hospital constructs module-specific experience representations consisting of refined executable prompts and structured modeling evidence. The federated server then performs evidence-guided experience aggregation to integrate reliable and complementary modeling experience across heterogeneous hospitals, distills the aggregated experience into global meta-prompts, and broadcasts them to guide subsequent local agent execution. Through iterative local refinement and federated experience aggregation, FedEHR-Agents enables distributed clinical agents to collaboratively improve automated EHR modeling while keeping patient-level EHR data local.

Our main contributions are summarized as follows:
\ding{182} We introduce an \emph{experience-centric federated optimization paradigm} for automated EHR modeling, reformulating cross-hospital collaboration from prediction model parameter optimization to the collaborative accumulation, refinement, and transfer of clinical modeling experience. This formulation extends FL beyond model-centric collaboration toward federated autonomous clinical intelligence.
\ding{183} We develop \textbf{FedEHR-Agents}, a federated agentic framework in which autonomous clinical agents perform end-to-end EHR modeling and continuously refine module-specific experience through historical memory, evaluator feedback, and TextGrad~\cite{yuksekgonul2024textgrad}-based prompt refinement. At the server, evidence-guided experience aggregation selectively integrates reliable and complementary modeling experience across heterogeneous hospitals and distills it into global meta-prompts for subsequent local refinement.
\ding{184} Extensive experiments on real-world multi-hospital EHR benchmarks validate the effectiveness of experience-centric federated optimization across diverse clinical prediction tasks and heterogeneous hospital settings. The results demonstrate that collaboratively optimizing clinical modeling experience enables transferable cross-hospital knowledge sharing beyond conventional model-centric federation, while maintaining robustness across different federation scales and LLM backbones.

\section{Problem Formulation} \label{sec:formulation}

\subsection{Federated Clinical Agentic Setting} \label{subsec:fed_cli_agentic}

We consider a federated clinical learning environment consisting of $K$ distributed hospitals $\mathcal{H}=\{H_1,\ldots,H_K\}$, where each hospital maintains a private longitudinal EHR dataset $\mathcal{D}_k=\{(\mathbf{x}_i,y_i)\}_{i=1}^{N_k}$. Each patient record $\mathbf{x}_i=(\mathbf{x}_i^{s},\mathbf{x}_i^{t})$ contains static patient characteristics $\mathbf{x}_i^{s}$ (e.g., demographics and admission information) and temporal clinical observations $\mathbf{x}_i^{t}$ (e.g., laboratory measurements, vital signs, medications, and respiratory interventions). The objective is to collaboratively improve automated EHR modeling for a shared clinical prediction task $\mathcal{T}$, such as in-hospital mortality prediction, while keeping all patient-level EHR data within their originating hospitals.

Unlike conventional FL~\cite{zhu2025fedweight}, where each hospital is represented primarily by a trainable prediction model, we associate each hospital with an autonomous clinical agent $\mathcal{A}_k^{(r)}$ that performs the complete EHR modeling workflow over local data at communication round $r$.
Formally, the local clinical agent is represented as
\begin{equation}
\mathcal{A}_k^{(r)}
=
\left(
\mathcal{G}_k,
\mathcal{X}_k^{(r)}
\right),
\label{eq:local_agent}
\end{equation}
where
$\mathcal{G}_k$
denotes the local execution engine responsible for automated data preprocessing, feature engineering, model development, evaluation, and iterative refinement, and
$\mathcal{X}_k^{(r)}$ denotes the clinical modeling experience accumulated by the agent up to communication round $r$.
The local agent performs one round of autonomous EHR modeling as
\begin{equation}
\left(
f_k^{(r)},
\widetilde{\mathcal{X}}_k^{(r)}
\right)
=
\mathcal{G}_k
\left(
\mathcal{D}_k,
\mathcal{T},
\mathcal{X}_k^{(r)}
\right),
\label{eq:local_agent_execution}
\end{equation}
where
$f_k^{(r)}$
denotes the prediction model produced during the current local execution, and
$\widetilde{\mathcal{X}}_k^{(r)}$
denotes the locally refined clinical modeling experience accumulated through data preprocessing, model development, execution feedback, historical memory, and prompt refinement.

Importantly, the prediction model $f_k^{(r)}$ is not the direct object of federated optimization, but rather an outcome of the local agent's reasoning, decision-making, and execution process. Instead, FedEHR-Agents collaboratively optimizes the underlying clinical modeling experience. During each communication round, local experience is progressively refined through execution feedback and historical memory, and further enhanced by the global meta-prompt distilled from experience aggregated across participating hospitals. Consequently, different hospitals may maintain heterogeneous prediction models while collaboratively benefiting from transferable clinical modeling experience.

\subsection{Clinical Modeling Experience}
\label{subsec:experience}

FedEHR-Agents treats \emph{clinical modeling experience} as the primary object of federated optimization. It represents transferable knowledge accumulated throughout automated EHR modeling, including preprocessing and feature engineering strategies, model development decisions, execution feedback, and historical optimization traces. 
Rather than being encoded in prediction model parameters, such experience evolves through local agent execution, while its shareable representation is collaboratively aggregated across distributed hospitals.

Formally, let $\mathcal{X}_k^{(r)}$ denote the clinical modeling experience maintained by agent $\mathcal{A}_k^{(r)}$ at communication round $r$. We represent it using three complementary components,
\begin{equation}
\mathcal{X}_k^{(r)}
=
\left(
\mathcal{S}_k^{(r)},
\mathcal{P}_k^{(r)},
\mathcal{M}_k^{(r)}
\right),
\label{eq:experience}
\end{equation}
where $\mathcal{S}_k^{(r)}$ denotes structured modeling evidence, $\mathcal{P}_k^{(r)}$
denotes the executable prompt representation, and $\mathcal{M}_k^{(r)}$
denotes the historical memory accumulated through previous agent executions.

Specifically, $\mathcal{S}_k^{(r)}$ captures explicit modeling outcomes, such as selected features, feature importance, model preferences, and evaluation statistics;
$\mathcal{P}_k^{(r)}$
organizes the accumulated experience into executable natural-language instructions that guide subsequent data preprocessing and model development; and
$\mathcal{M}_k^{(r)}$
preserves historical execution information, including modeling decisions, evaluator feedback, execution records, and previous refinement trajectories.
While all three components remain local as part of the clinical modeling experience, only the refined prompt and structured modeling evidence are communicated during federated aggregation.

\subsection{Federated Experience Optimization}
\label{subsec:fedagent_opti}

We formulate federated automated EHR modeling as an iterative optimization process over \emph{clinical modeling experience}. Unlike conventional FL, which optimizes prediction models through parameter aggregation, FedEHR-Agents collaboratively improves the evolving experience maintained by autonomous clinical agents. At each communication round, local agents first refine their experience through autonomous EHR modeling and execution feedback, after which the resulting experience is aggregated across hospitals and redistributed as globally transferable guidance.

After local agent execution, each hospital constructs a shareable experience representation $\mathcal{R}_k^{(r)}$
from its locally refined prompt and structured modeling evidence. These representations are uploaded to the federated server and integrated
through an evidence-guided experience aggregation operator,
\begin{equation}
\mathcal{R}_G^{(r)}
=
\Psi
\left(
\mathcal{R}_1^{(r)},
\ldots,
\mathcal{R}_K^{(r)}
\right),
\label{eq:global_experience}
\end{equation}
where
$\Psi(\cdot)$
denotes the federated experience aggregation function and $\mathcal{R}_G^{(r)}$
represents the globally aggregated clinical modeling experience. Rather than averaging model parameters, $\Psi(\cdot)$ integrates transferable modeling strategies and supporting evidence accumulated across heterogeneous hospitals.

To translate the aggregated clinical modeling experience into actionable guidance for local clinical agents, the federated server further synthesizes a global meta-prompt,
\begin{equation}
\mathcal{P}_G^{(r)}
=
\Gamma
\left(
\mathcal{R}_G^{(r)}
\right),
\label{eq:meta_prompt}
\end{equation}
where
$\Gamma(\cdot)$
denotes the experience-guided meta-prompt generation function.
The generated meta-prompt distills globally transferable preprocessing strategies, feature engineering practices, model development experience, and optimization recommendations from the global experience representation.

Each hospital subsequently integrates the global meta-prompt with its locally refined experience to construct the experience state for the next communication round,
\begin{equation}
\mathcal{X}_k^{(r+1)}
=
\Phi
\left(
\widetilde{\mathcal{X}}_k^{(r)},
\mathcal{P}_G^{(r)}
\right),
\label{eq:experience_update}
\end{equation}
where
$\Phi(\cdot)$
denotes the local experience integration operator.
This update preserves hospital-specific modeling experience while incorporating transferable knowledge distilled from other participating hospitals, thereby enabling continual improvement of local autonomous EHR modeling.

The overall objective is to maximize the downstream modeling utility induced by the clinical modeling experience maintained across participating hospitals. Let
$N=\sum_{k=1}^{K}N_k$.
We formulate the federated experience optimization objective as
\begin{equation}
\left\{
\mathcal{X}_k^{*}
\right\}_{k=1}^{K}
=
\arg\max_{\{\mathcal{X}_k\}_{k=1}^{K}}
\sum_{k=1}^{K}
\frac{N_k}{N}
\,
\mathcal{U}_k
\left(
\mathcal{X}_k;
\mathcal{D}_k,
\mathcal{T}
\right),
\label{eq:optimization}
\end{equation}
where
$\mathcal{U}_k(\cdot)$
denotes the task-specific utility induced by the clinical modeling experience at hospital $k$, measured through the quality of the resulting automated EHR modeling workflow and its downstream predictive performance.
The objective is realized iteratively through local experience refinement, federated experience aggregation, global meta-prompt generation, and experience integration, rather than through direct optimization of prediction models.

\begin{figure*}[!t]
\centering
\setlength{\abovecaptionskip}{0.1cm}
\includegraphics[width=0.9\linewidth]{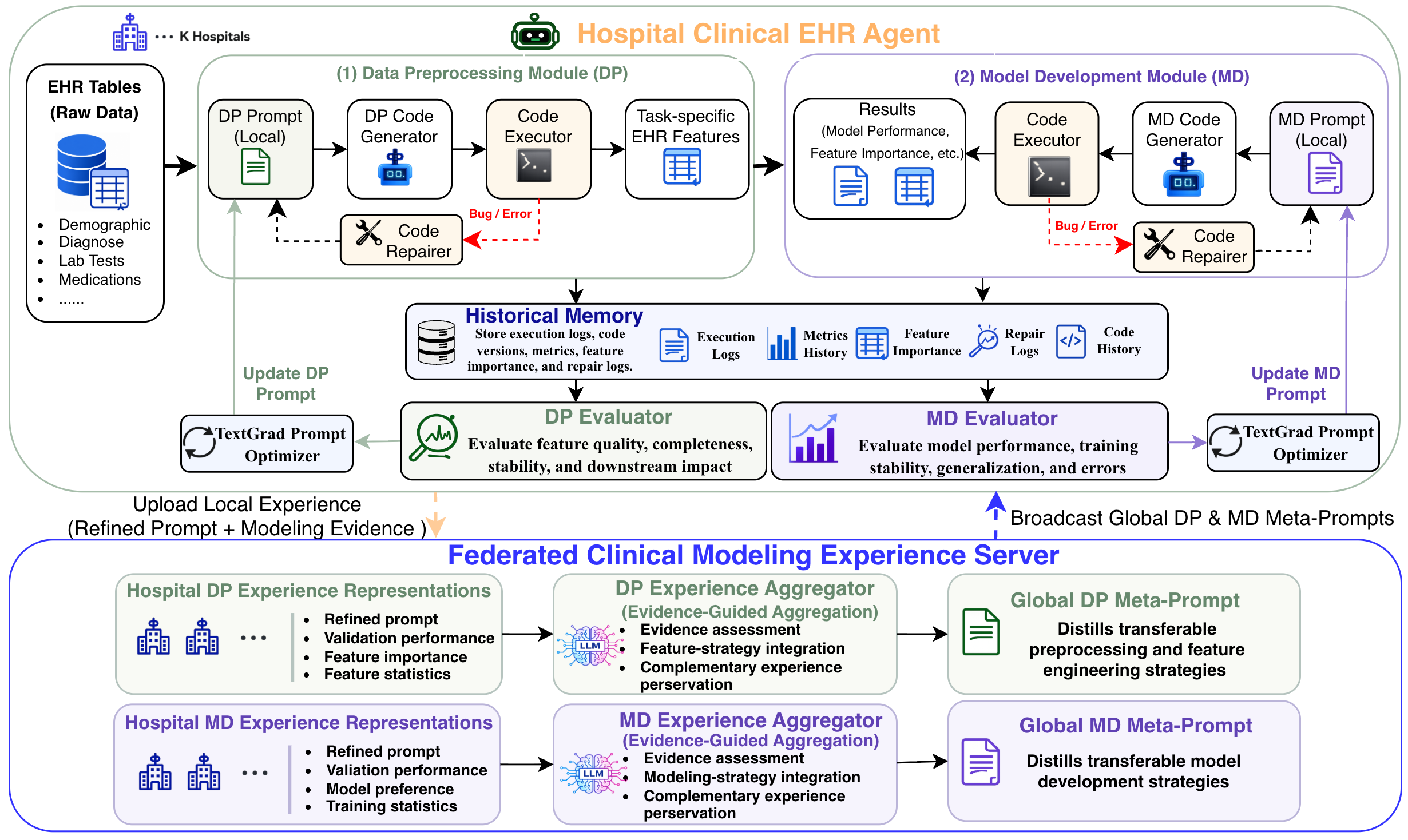}
\caption{
Overview of the proposed FedEHR-Agents framework. Each hospital deploys an autonomous clinical EHR agent to perform data preprocessing (DP) and model development (MD) while continuously refining local clinical modeling experience. The federated server aggregates module-specific experience representations through evidence-guided experience aggregation, synthesizes global DP and MD meta-prompts, and broadcasts them to guide subsequent local experience refinement.
}
\Description{.}
\label{fig:fedehr-agents}
\vspace{-0.3cm}
\end{figure*}

\section{Methodology}  \label{sec:method}

\subsection{Overview}
\label{sec:overview}

The proposed FedEHR-Agents framework consists of distributed hospital clinical EHR agents and a federated server that collaboratively optimize clinical modeling experience rather than prediction model parameters (Fig.~\ref{fig:fedehr-agents}). Each hospital deploys an autonomous clinical EHR agent to perform automated EHR modeling over private patient data, while the federated server integrates locally refined modeling experience across hospitals and synthesizes global meta-prompts to guide subsequent local experience refinement.

Specifically, each clinical EHR agent consists of a \emph{Data Preprocessing (DP) Module} and a \emph{Model Development (MD) Module}, supported by historical memory, task-specific evaluators, and TextGrad-based prompt refinement~\cite{yuksekgonul2024textgrad}. During each communication round, the agent executes the DP and MD workflows, evaluates their outputs, updates historical memory, and refines the corresponding executable prompts according to execution feedback. The resulting module-specific experience representations are then uploaded to the federated server, where DP and MD experiences are separately integrated through evidence-guided experience aggregation. The server subsequently synthesizes global DP and MD meta-prompts and broadcasts them to participating hospitals to guide the next round of local experience refinement and agent execution. Through this iterative process, FedEHR-Agents progressively improves automated EHR modeling while keeping patient-level EHR data local.

\subsection{Local Hospital Clinical EHR Agent}
\label{sec:local_agent}

Each participating hospital deploys an autonomous clinical EHR agent to perform automated EHR modeling over its private EHR database \cite{wang2026doctoragents}. The clinical agent consists of two functional modules: a DP Module and an MD Module. The DP module transforms heterogeneous raw EHR tables into structured task-specific clinical features, while the MD module develops prediction models based on the resulting feature representations. Both modules are further supported by historical memory, task-specific evaluators, and TextGrad-based prompt refinement, allowing the agent to continuously accumulate and refine clinical modeling experience throughout federated optimization.

\subsubsection{Data Preprocessing Module}
\label{sec:dp_agent}

The DP module transforms heterogeneous EHR tables into task-specific feature representations suitable for downstream clinical prediction. Given the local EHR database
$\mathcal{D}_k$
and the preprocessing prompt
$\mathcal{P}_{k,\mathrm{DP}}^{(r)}$,
the DP Code Generator employs an LLM to generate executable preprocessing code for data extraction, cleaning, temporal alignment, feature engineering, and feature selection. The generated code is then executed locally by the Code Executor. If execution errors are detected, the Code Repairer analyzes the error messages and revises the generated code until a valid preprocessing workflow is obtained. The resulting feature representation is formulated as
\begin{equation}
\mathbf{Z}_{k}^{(r)}
=
g_{\mathrm{DP}}
\left(
\mathcal{D}_k,
\mathcal{P}_{k,\mathrm{DP}}^{(r)}
\right),
\label{eq:dp}
\end{equation}
where
$g_{\mathrm{DP}}(\cdot)$
denotes the complete LLM-guided preprocessing workflow, including code generation, local execution, and error-driven code repair.

The resulting representation
$\mathbf{Z}_{k}^{(r)}$
is subsequently passed to the model development module for downstream prediction. Meanwhile, selected features and feature statistics are summarized as structured
modeling evidence for subsequent federated aggregation, while generated code and execution records are retained in the historical memory for local evaluation and prompt refinement.

\subsubsection{Model Development Module}
\label{sec:md_agent}

Based on the processed feature representation
$\mathbf{Z}_{k}^{(r)}$,
the MD module autonomously develops a prediction model for the target clinical task.
Given the model development prompt $\mathcal{P}_{k,\mathrm{MD}}^{(r)}$,
the MD Code Generator employs an LLM to determine an appropriate modeling strategy, including model selection, training configuration, and hyperparameter settings, and generates executable model development code. The generated code is executed locally by the Code Executor, while execution failures are handled by the Code Repairer through error-guided code revision. The resulting local prediction model is formulated as
\begin{equation}
f_k^{(r)}
=
g_{\mathrm{MD}}
\left(
\mathbf{Z}_{k}^{(r)},
\mathbf{y}_k,
\mathcal{P}_{k,\mathrm{MD}}^{(r)}
\right),
\label{eq:md}
\end{equation}
where
$g_{\mathrm{MD}}(\cdot)$
denotes the complete LLM-guided model development workflow, including code generation, local execution, and error-driven code repair, and $\mathbf{y}_k$ denotes the corresponding prediction labels.

In addition to the resulting prediction model, the MD module produces modeling outcomes such as model preferences, validation performance, feature importance, training statistics, and execution records. Validation performance, feature importance, model preferences, and training statistics are summarized as structured modeling evidence for server-side experience aggregation, while execution records and refinement histories
are retained in the local historical memory for prompt refinement.

\subsection{Historical Memory and Evaluation}
\label{sec:memory}

To support iterative refinement across communication rounds, each local clinical EHR agent maintains a historical memory and employs task-specific evaluators to assess the current execution. The historical memory records previous preprocessing and model development artifacts, execution logs, evaluation results, and refinement histories, allowing the agent to reuse effective modeling decisions rather than restarting from scratch. The evaluators assess the outputs of the DP and MD modules and generate structured feedback on their quality and performance. Together, historical context and evaluation feedback are used to refine local prompts and update the clinical modeling experience in subsequent communication rounds.

\subsubsection{Historical Memory}

During each communication round, the clinical EHR agent records intermediate artifacts generated during local execution, including preprocessing programs, model development configurations, evaluation results, feature importance, execution logs, and prompt refinement histories. Formally, the historical memory maintained by hospital $k$ at round $r$ is represented as
\begin{equation}
\mathcal{M}_{k}^{(r)}
=
\left\{
\mathbf{c}_{k}^{(r)},
\mathbf{e}_{k}^{(r)},
\mathbf{h}_{k}^{(r)}
\right\},
\label{eq:memory}
\end{equation}
where
$\mathbf{c}_{k}^{(r)}$
contains the generated executable code and associated modeling artifacts,
$\mathbf{e}_{k}^{(r)}$ records execution logs and evaluation results,
and $\mathbf{h}_{k}^{(r)}$ stores the prompt refinement trajectories accumulated across previous communication rounds.

The historical memory is updated after each local execution and provides long-term context for subsequent evaluation and prompt refinement. It remains local to the hospital throughout federated optimization and
is not transmitted to the federated server.

\subsubsection{Module Evaluation}

Following each local execution, the outputs of the DP and MD modules are assessed by task-specific evaluators. The DP evaluator examines the generated feature representations with respect to feature completeness, consistency, temporal validity, potential information leakage, and downstream usability. The MD evaluator assesses the resulting prediction model in terms of predictive performance, generalization, optimization stability, and model reliability. Rather than directly modifying the generated code or prediction model, both evaluators produce structured textual feedback for subsequent prompt refinement.

Formally, for module
$m\in\{\mathrm{DP},\mathrm{MD}\}$,
the evaluation process is defined as
\begin{equation}
\mathcal{F}_{k,m}^{(r)}
=
\mathcal{E}_{m}
\left(
\mathcal{O}_{k,m}^{(r)},
\mathcal{M}_{k}^{(r)}
\right),
\label{eq:evaluation}
\end{equation}
where
$\mathcal{O}_{k,m}^{(r)}$
denotes the current execution outputs and associated statistics of module $m$,
$\mathcal{E}_{m}(\cdot)$
denotes the corresponding task-specific evaluator, and
$\mathcal{F}_{k,m}^{(r)}$
is the generated textual feedback.
The historical memory
$\mathcal{M}_{k}^{(r)}$
provides additional context for comparing the current execution with previous optimization rounds.

The resulting feedback is recorded in the historical memory and subsequently used by TextGrad to refine the corresponding DP and MD prompts.

\subsection{Local Prompt Refinement}
\label{sec:local_prompt}

Following module evaluation, the generated feedback is used to refine the executable prompts that guide subsequent agent execution. In FedEHR-Agents, prompts serve as executable representations of clinical modeling experience rather than fixed manually designed instructions. By revising these prompts according to current evaluation feedback and historical execution context, the local agent progressively incorporates newly acquired modeling experience into subsequent executions.

For each module
$m\in\{\mathrm{DP},\mathrm{MD}\}$,
let
$\mathcal{P}_{k,m}^{(r)}$
denote the corresponding local prompt at communication round $r$.
Given the evaluator feedback
$\mathcal{F}_{k,m}^{(r)}$
and historical memory
$\mathcal{M}_{k}^{(r)}$,
the prompt is refined as
\begin{equation}
\widetilde{\mathcal{P}}_{k,m}^{(r)}
=
\Omega_m
\left(
\mathcal{P}_{k,m}^{(r)},
\mathcal{F}_{k,m}^{(r)},
\mathcal{M}_{k}^{(r)}
\right),
\label{eq:local_prompt}
\end{equation}
where
$\Omega_m(\cdot)$
denotes the prompt refinement operator for module $m$.
We implement
$\Omega_m(\cdot)$
using TextGrad~\cite{yuksekgonul2024textgrad}, which interprets evaluator feedback as textual gradients and revises the prompt accordingly without gradient back-propagation over model parameters.

The refined prompt
$\widetilde{\mathcal{P}}_{k,m}^{(r)}$
captures the locally updated modeling experience and is subsequently included in the experience representation uploaded for federated aggregation. After incorporating the corresponding global meta-prompt, it is further updated to form the executable prompt for the next communication round.

\subsection{Evidence-Guided Experience Aggregation}
\label{sec:federated_aggregation}

FedEHR-Agents performs server-side collaboration through evidence-guided experience aggregation (EGEA). Each hospital uploads its locally refined experience representation, consisting of the refined executable prompt and structured modeling evidence. Here, structured modeling evidence refers to measurable outcomes obtained from local agent execution, such as validation performance, feature importance, feature statistics, model preferences, and optimization trends. EGEA uses this evidence to assess the reliability and transferability of local modeling experience and integrates complementary experience across heterogeneous hospitals. The aggregated experience is then distilled into global meta-prompts to guide the next round of agent execution.

\subsubsection{Experience Aggregation}

For each module
$m\in\{\mathrm{DP},\mathrm{MD}\}$,
hospital $k$ constructs a shareable local experience representation consisting of the locally refined prompt and structured modeling evidence,
\begin{equation}
\mathcal{R}_{k,m}^{(r)}
=
\left(
\widetilde{\mathcal{P}}_{k,m}^{(r)},
\mathcal{S}_{k,m}^{(r)}
\right),
\label{eq:uploaded_experience}
\end{equation}
where
$\widetilde{\mathcal{P}}_{k,m}^{(r)}$
denotes the locally refined executable prompt and
$\mathcal{S}_{k,m}^{(r)}$
contains structured modeling evidence obtained from local execution.
The historical memory
$\mathcal{M}_{k}^{(r)}$
is retained locally and is used only to provide longitudinal context for evaluation and prompt refinement; it is not directly transmitted to the federated server.

The federated server separately aggregates the DP and MD experience representations across participating hospitals,
\begin{equation}
\mathcal{R}_{G,m}^{(r)}
=
\Psi_m
\left(
\mathcal{R}_{1,m}^{(r)},
\ldots,
\mathcal{R}_{K,m}^{(r)}
\right),
\quad
m\in\{\mathrm{DP},\mathrm{MD}\},
\label{eq:experience_aggregation}
\end{equation}
where
$\Psi_m(\cdot)$
denotes the corresponding evidence-guided experience aggregation operator.
During aggregation, the structured modeling evidence
$\mathcal{S}_{k,m}^{(r)}$
is used to assess the effectiveness and reliability of the associated local experience, while complementary strategies from heterogeneous hospitals are preserved when constructing the global experience representation
$\mathcal{R}_{G,m}^{(r)}$.

\subsubsection{Global Meta-Prompt Generation}

To translate the aggregated experience into actionable guidance for local clinical agents, the federated server generates a global meta-prompt for each module,
\begin{equation}
\mathcal{P}_{G,m}^{(r)}
=
\Gamma_m
\left(
\mathcal{R}_{G,m}^{(r)}
\right),
\quad
m\in\{\mathrm{DP},\mathrm{MD}\},
\label{eq:meta_prompt_generation}
\end{equation}
where
$\Gamma_m(\cdot)$
denotes the LLM-based meta-prompt generation function for module $m$.
The generated meta-prompts distill transferable modeling strategies and recommendations from the aggregated DP and MD experience representations.

The global meta-prompts are broadcast to participating hospitals and integrated with the corresponding locally refined prompts,
\begin{equation}
\mathcal{P}_{k,m}^{(r+1)}
=
\Phi_m
\left(
\widetilde{\mathcal{P}}_{k,m}^{(r)},
\mathcal{P}_{G,m}^{(r)}
\right),
\quad
m\in\{\mathrm{DP},\mathrm{MD}\},
\label{eq:prompt_update}
\end{equation}
where
$\Phi_m(\cdot)$
denotes the prompt integration operator. The resulting prompts combine locally accumulated experience with globally shared guidance and are used to initialize the DP and MD modules in the next communication round. The complete federated optimization procedure is summarized in Appendix Algorithm~\ref{alg:fedehr_agents}. Detailed prompt templates and representative structured
outputs are provided in Appendix~Sec.~\ref{appendix:prompts}.

\begin{table*}[!t]
\centering
\caption{Overall AUPRC comparison across four representative clinical prediction tasks.
Higher values indicate better performance.
The best results among the independent and federated methods are highlighted in bold.
Centralized methods are reported only as upper-bound references.}
\label{tab:overall}
\setlength{\tabcolsep}{11pt}
\renewcommand{\arraystretch}{1.12}
\begin{tabular}{lcccc}
\toprule
\textbf{Method}
& \textbf{Mortality-48h}
& \textbf{ARF-4h}
& \textbf{LOS$>$7d}
& \textbf{Sepsis} \\
\midrule

\rowcolor{gray!8}
\multicolumn{5}{l}{\textsc{Centralized Methods (Upper Bound)}} \\
Centralized AutoML
& $0.232 \pm 0.005$
& $0.189 \pm 0.006$
& $0.543 \pm 0.014$
& $0.292 \pm 0.007$ \\

Centralized Agent
& $0.242 \pm 0.004$
& $0.204 \pm 0.005$
& $0.578 \pm 0.011$
& $0.337 \pm 0.008$ \\

\addlinespace[3pt]

\rowcolor{gray!8}
\multicolumn{5}{l}{\textsc{Independent Methods}} \\

Local AutoML
& $0.205 \pm 0.007$
& $0.158 \pm 0.007$
& $0.465 \pm 0.017$
& $0.256 \pm 0.005$ \\

Local Agent
& $0.218 \pm 0.005$
& $0.171 \pm 0.005$
& $0.483 \pm 0.015$
& $0.268 \pm 0.007$ \\

\addlinespace[3pt]

\rowcolor{gray!8}
\multicolumn{5}{l}{\textsc{Federated Methods}} \\

FedEHR-Agents (PromptAvg)
& $0.221 \pm 0.006$
& $0.183 \pm 0.006$
& $0.506 \pm 0.012$
& $0.286 \pm 0.011$ \\

\textbf{FedEHR-Agents}
& \textbf{$0.236 \pm 0.005$}
& \textbf{$0.194 \pm 0.004$}
& \textbf{$0.532 \pm 0.010$}
& \textbf{$0.316 \pm 0.009$} \\

\bottomrule
\end{tabular}
\end{table*}

\section{Evaluation}
\label{sec:experiments}

\subsection{Experimental Setup}
\label{sec:setup}

\paragraph{Datasets and Clinical Prediction Tasks.}
We evaluate FedEHR-Agents on the publicly available eICU Collaborative Research Database (eICU) \cite{pollard2018eicu}, a large-scale multi-center critical care dataset collected from more than 200 intensive care units across the United States. Each participating hospital is treated as an independent federated client, with patient-level EHR data retained locally throughout the entire optimization process. We consider four representative clinical prediction tasks: 48-hour mortality prediction (Mortality-48h), 4-hour acute respiratory failure prediction (ARF-4h), prolonged length-of-stay prediction (LOS$>$7d), and sepsis prediction (Sepsis). These tasks cover diverse clinical outcomes and prediction horizons, providing a comprehensive evaluation of federated automated EHR modeling under heterogeneous multi-hospital settings.

\paragraph{Federated Configuration.}
We adopt a cross-silo federated setting in which participating hospitals collaboratively optimize clinical modeling experience without exchanging patient-level EHR data or locally developed prediction models. At each communication round, local clinical agents perform EHR modeling and prompt refinement, after which the resulting DP and MD experience representations are aggregated by the federated server through EGEA. The server then generates global DP and MD meta-prompts to guide the next round of local agent execution. Unless otherwise specified, we use three participating hospitals and five communication rounds for all experiments.
All experiments are repeated with three random seeds, and the mean and standard deviation are reported.

\paragraph{Baseline Methods.}
We compare FedEHR-Agents with five representative automated EHR modeling baselines covering local, federated, and centralized settings. \textbf{Local AutoML} independently develops prediction models at each hospital using a conventional AutoML framework~\cite{trirat2025automlagent}. \textbf{Local Agent} uses the same clinical agent architecture as FedEHR-Agents but performs all modeling and prompt refinement locally without cross-hospital collaboration. \textbf{FedEHR-Agents (PromptAvg)} provides a simple federated prompt-sharing baseline, where local prompt representations are aggregated to construct global meta-prompts, without incorporating structured modeling evidence during server-side aggregation. We further include \textbf{Centralized AutoML} and \textbf{Centralized Agent} as upper-bound references, where data from all participating hospitals are pooled for automated model development. Since these centralized settings violate the federated data-locality assumption, they are reported only as oracle performance references.

\paragraph{Implementation Details.}
All agent-based methods use GPT-5 mini as the default LLM backbone. Each participating hospital deploys the same clinical agent architecture, consisting of a DP module and an MD module, while allowing preprocessing workflows, feature engineering strategies, prediction models, and modeling decisions to adapt autonomously to local EHR data. For fair comparison, all methods use the same communication protocol, optimization budget, evaluation metrics, and stopping criteria. 
Additional implementation details are provided in Appendix~Sec.~\ref{appendix:data_configuration}.

\paragraph{Evaluation Metrics.}
Prediction performance is evaluated using the Area Under the Precision--Recall Curve (AUPRC), which is well suited to imbalanced clinical prediction tasks. Beyond predictive performance, we evaluate the behavior of federated experience optimization through convergence across communication rounds, cross-hospital feature selection consistency, and SHAP-based feature attribution consistency. We further examine the robustness of FedEHR-Agents under different federation scales and LLM backbones. Together, these evaluations assess both the predictive effectiveness and the optimization stability of the proposed framework.

\subsection{Overall Performance}
\label{sec:overall}

We first evaluate FedEHR-Agents on four representative clinical prediction tasks: Mortality-48h, ARF-4h, LOS$>$7d, and Sepsis. FedEHR-Agents consistently achieves the best performance among all decentralized methods across all tasks (Table~\ref{tab:overall}). Compared with Local Agent, the AUPRC increases from 0.218 to 0.236 on Mortality-48h, from 0.171 to 0.194 on ARF-4h, from 0.483 to 0.532 on LOS$>$7d, and from 0.268 to 0.316 on Sepsis, corresponding to relative gains of 8.3\%, 13.5\%, 10.1\%, and 17.9\%, respectively. The consistent gains over Local AutoML and Local Agent indicate that cross-hospital experience sharing enables autonomous clinical agents to benefit from transferable modeling knowledge beyond isolated local optimization. Moreover, FedEHR-Agents approaches the centralized upper-bound performance without pooling patient-level EHR data, demonstrating that collaborative experience optimization can recover a substantial portion of the benefit of centralized learning.

FedEHR-Agents also consistently outperforms \textbf{FedEHR-Agents (PromptAvg)}, which constructs global meta-prompts from aggregated local prompt representations without structured modeling evidence. The absolute AUPRC improvements are 0.015, 0.011, 0.026, and 0.030 on Mortality-48h, ARF-4h, LOS$>$7d, and Sepsis, respectively (Table~\ref{tab:overall}). The larger gains on LOS$>$7d and Sepsis suggest that structured modeling evidence provides useful context for identifying more reliable and transferable local experience during server-side aggregation. As a result, the generated global meta-prompts provide more effective guidance for subsequent local refinement than prompt aggregation alone.

\subsection{Federated Experience Optimization}
\label{sec:optimization}

\begin{figure*}[!ht]
    \centering
    \includegraphics[width=\textwidth]{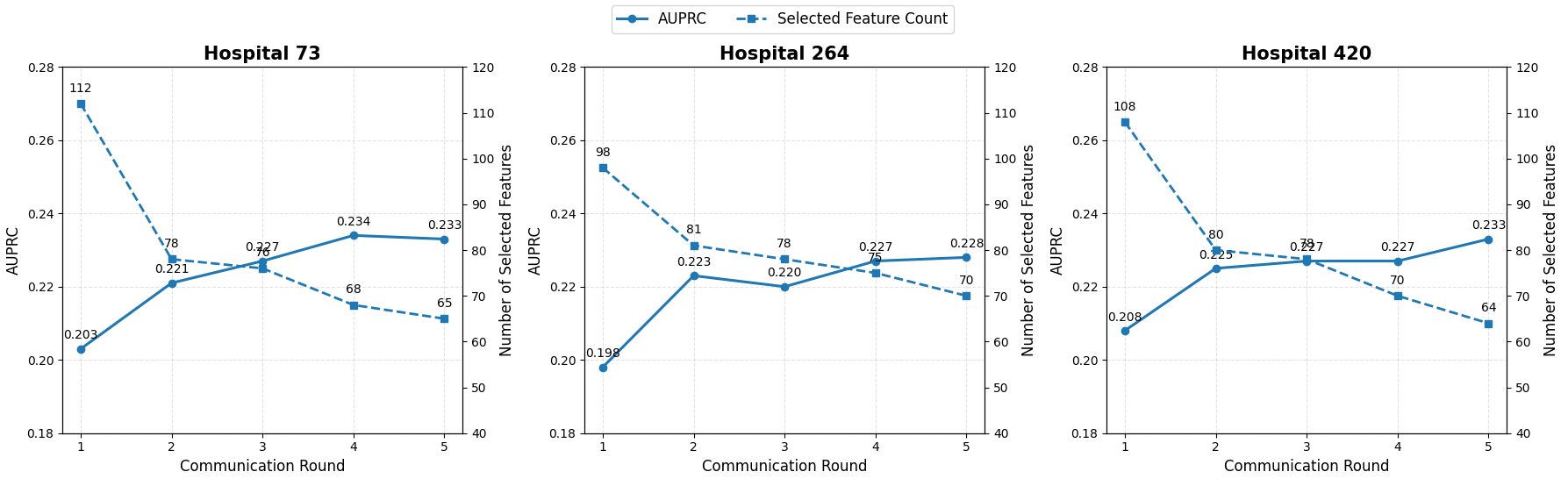}
  \caption{
    Optimization trajectories of FedEHR-Agents on the Mortality-48h task across three hospitals. Solid lines denote performance measured by AUPRC, while dashed lines denote the number of selected  features over communication rounds.
    }
    \Description{.}
    \label{fig:optimization_process}
\end{figure*}

\begin{figure}[!t]
    \centering
    \includegraphics[width=0.48\textwidth]{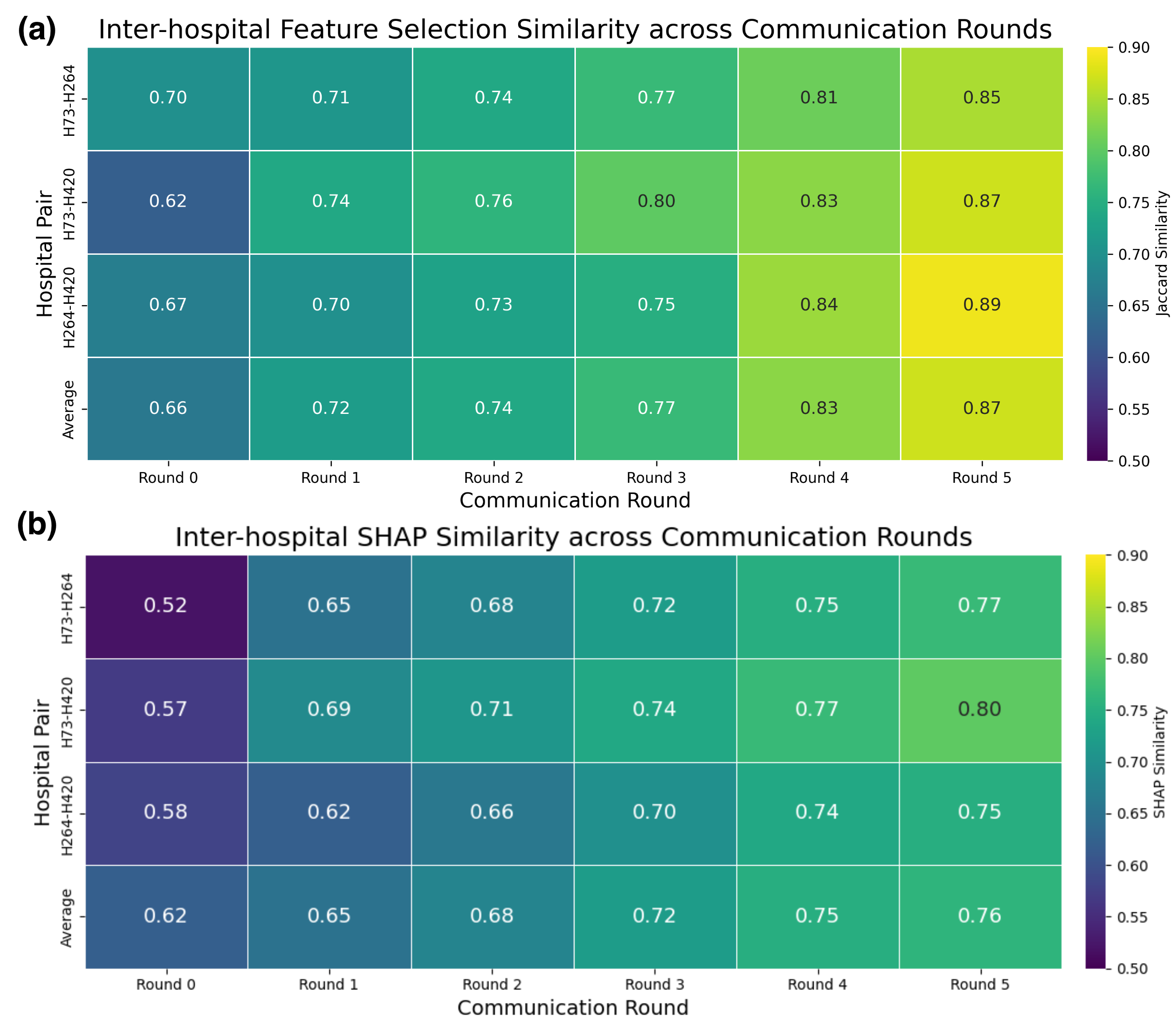}
    \caption{
    Evolution of cross-hospital modeling consistency on the Mortality-48h task. (a) Pairwise feature selection consistency measured by Jaccard similarity. (b) Pairwise feature importance consistency measured by SHAP similarity.
    }
    \Description{.}
    \label{fig:modeling_consistency}
    \vspace{-1.5em}
\end{figure}

To examine how FedEHR-Agents evolves over rounds, we analyze the federated experience optimization process on the Mortality-48h task. We track local predictive performance, the number of selected features, and cross-hospital consistency in feature selection and feature attribution, providing a more detailed view of how collaborative experience sharing influences local EHR modeling.

Local predictive performance improves rapidly during the early communication rounds and gradually stabilizes across all participating hospitals (Fig.~\ref{fig:optimization_process}). At the same time, the number of selected clinical features decreases steadily, yielding increasingly compact feature subsets. Notably, predictive performance continues to improve despite the reduced feature space, suggesting that the agents progressively favor more informative features while discarding less useful ones. This trend indicates that the accumulated modeling experience improves both feature selection and downstream model development over successive rounds.

Cross-hospital modeling decisions also become increasingly consistent during federated optimization (Fig.~\ref{fig:modeling_consistency}). The pairwise Jaccard similarity between selected feature subsets rises steadily across communication rounds, indicating greater agreement in feature selection among independently operating hospital agents. A similar pattern is observed for SHAP-based feature importance, where cross-hospital agreement also increases despite heterogeneous local EHR distributions (Fig.~\ref{fig:modeling_consistency}b). These results suggest that federated experience optimization not only improves local predictive performance, but also promotes the emergence of more consistent and reusable modeling knowledge across distributed hospitals.

\subsection{Clinical Feature Interpretation}
\label{sec:clinical}

\begin{figure*}[t]
    \centering
    \includegraphics[width=\textwidth]{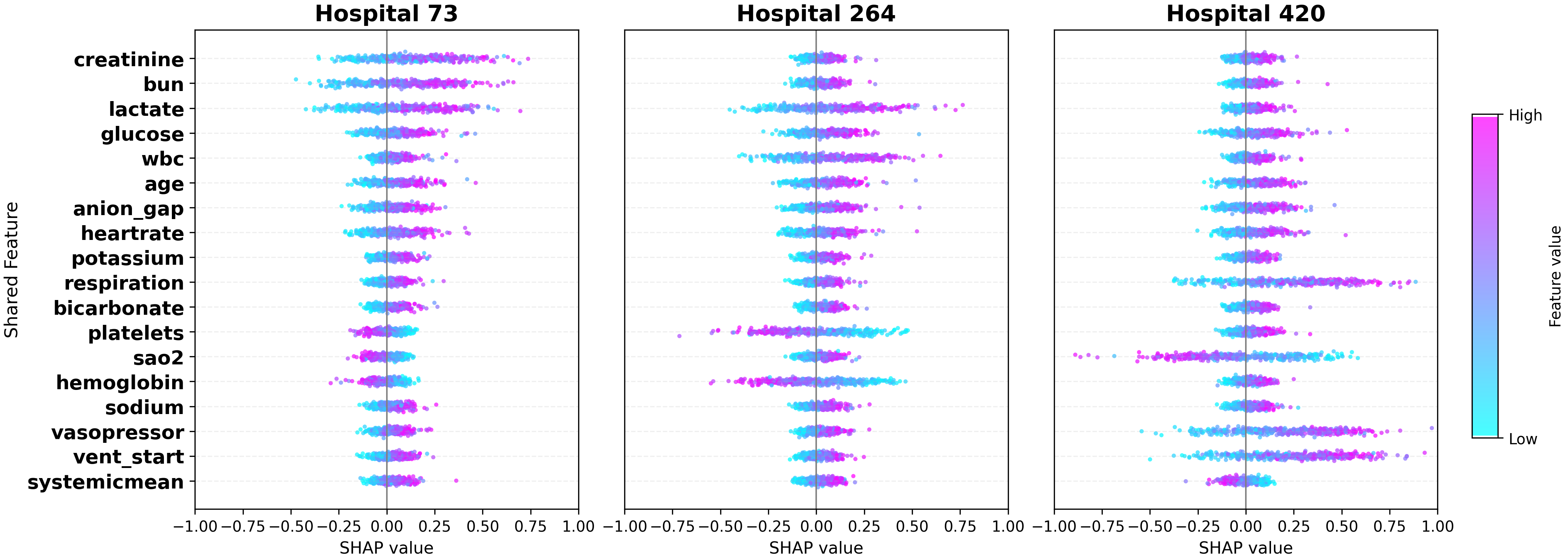}
    \caption{
Hospital-level SHAP analysis of the shared clinically important features on the Mortality-48h task. The figure presents the SHAP distributions of the common top-ranked features identified across three hospitals after federated optimization. For readability, the feature names shown in Fig.~\ref{fig:shap_analysis} are normalized labels used after preprocessing rather
than the original eICU variable names; detailed definitions are provided
in Appendix~Table~\ref{tab:mortality_feature_definitions}.
}
\Description{.}
    \label{fig:shap_analysis}
\end{figure*}

To complement the quantitative evaluation, we further investigate the clinical interpretability of the optimized policies by analyzing the SHAP values of the overlapping top-ranked features identified across three representative hospitals on the Mortality-48h prediction task (Fig.~\ref{fig:shap_analysis}). Specifically, we examine whether federated optimization enables autonomous agents to consistently identify clinically meaningful predictors while preserving hospital-specific modeling characteristics.

Across all participating hospitals, FedEHR-Agents consistently identifies clinically established mortality predictors, including renal function indicators, metabolic measurements, inflammatory biomarkers, and respiratory-related variables (Fig.~\ref{fig:shap_analysis}). Although these clinically important features are consistently selected, their SHAP value distributions differ across hospitals, reflecting variations in patient populations, disease prevalence, and local clinical practice. This suggests that federated optimization successfully captures globally transferable clinical knowledge while preserving hospital-specific decision patterns, enabling collaborative yet personalized automated EHR modeling.

\subsection{Ablation Study}
\label{sec:ablation}

We conduct an ablation study on the Mortality-48h task to evaluate the contribution of four key components in FedEHR-Agents: Historical Memory, Evaluator, DP Module, and EGEA (Table~\ref{tab:ablation}). For each ablation setting, one component is removed while all remaining components are kept unchanged. Historical Memory preserves prior execution and refinement information, the Evaluator provides feedback for local prompt refinement, the DP Module supports autonomous preprocessing and feature engineering, and EGEA integrates local experience using structured modeling evidence.

Removing any component consistently reduces predictive performance (Table~\ref{tab:ablation}). Disabling the DP Module causes the largest degradation, decreasing AUPRC from 0.236 to 0.219, highlighting the importance of adaptive preprocessing and feature engineering. Replacing EGEA with simple prompt averaging reduces AUPRC to 0.221, showing that structured modeling evidence improves server-side experience aggregation beyond prompt sharing alone. Removing Historical Memory and the Evaluator decreases AUPRC to 0.223 and 0.226, respectively, indicating that historical context and evaluator feedback both contribute to effective local prompt refinement. Additional ablation results across clinical prediction tasks are provided in Appendix~Table~\ref{tab:additional_ablation}. Overall, the results confirm that the four components provide complementary benefits to federated experience optimization.

\begin{table}[t]
\centering
\caption{Ablation Study of FedEHR-Agents on the Mortality-48h prediction task.}
\label{tab:ablation}

\renewcommand{\arraystretch}{1.1}
\setlength{\tabcolsep}{4pt}

\begin{tabular}{cccc|c}
\toprule

\textbf{Memory} &
\textbf{Evaluator} &
\textbf{DP Module} &
\textbf{EGEA} &
\textbf{AUPRC} \\
\midrule

$\times$     & $\checkmark$ & $\checkmark$ & $\checkmark$
& $0.223 \pm 0.004$ \\

$\checkmark$ & $\times$     & $\checkmark$ & $\checkmark$
& $0.226 \pm 0.006$ \\

$\checkmark$ & $\checkmark$ & $\times$     & $\checkmark$
& $0.219 \pm 0.007$ \\

$\checkmark$ & $\checkmark$ & $\checkmark$ & $\times$
& $0.221 \pm 0.006$ \\

\midrule

$\checkmark$ & $\checkmark$ & $\checkmark$ & $\checkmark$
& $\mathbf{0.236 \pm 0.005}$ \\

\bottomrule
\end{tabular}
\end{table}

\subsection{Scalability Analysis}
\label{sec:scalability}

\begin{figure}[t]
    \centering
    \includegraphics[width=\columnwidth]
    {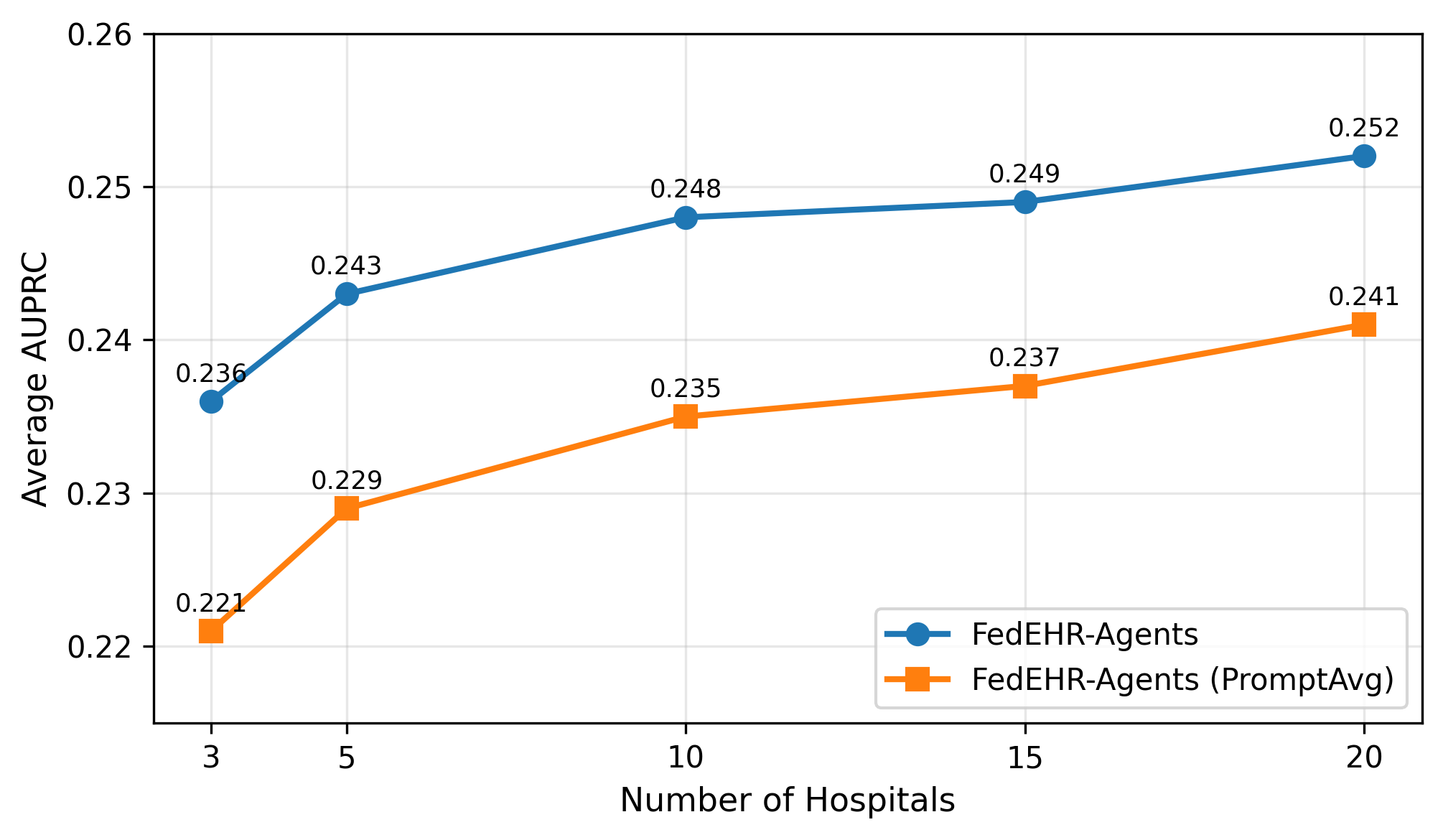}
    \caption{
    Impact of the number of participating hospitals on Mortality-48h prediction. Average AUPRC is reported for federations with 3, 5, 10, 15, and 20 hospitals.
    }
    \Description{.}
    \label{fig:hospital_scalability}
    \vspace{-1.5em}
\end{figure}

We evaluate the scalability of FedEHR-Agents by varying the number of participating hospitals from 3 to 20 on the Mortality-48h task. For each federation scale, the average AUPRC is computed across all participating hospitals. We compare the complete FedEHR-Agents framework with the PromptAvg variant to examine whether EGEA remains effective as the federation size increases.

Both methods benefit from larger federations, suggesting that additional hospitals provide more diverse modeling experience for collaborative optimization (Fig.~\ref{fig:hospital_scalability}). FedEHR-Agents consistently outperforms PromptAvg across all federation scales, with the average AUPRC increasing from 0.236 to 0.252 as the number of hospitals grows from 3 to 20, compared with 0.221 to 0.241 for PromptAvg. The persistent performance gap indicates that prompt aggregation alone is less effective at exploiting heterogeneous cross-hospital experience as the federation expands. In contrast, EGEA leverages structured modeling evidence to integrate more reliable and complementary experience across hospitals, supporting stable performance gains under larger multi-hospital federations. Consistent scalability trends are also observed across the other clinical prediction tasks (Appendix~Table~\ref{tab:additional_scalability}).

\subsection{Impact of Different LLM Backbones}
\label{subsec:varying_llms}

We evaluate the robustness of FedEHR-Agents to different LLM backbones by comparing GPT-4o, GPT-5 mini, and GPT-5. For each backbone, we use the same federated optimization pipeline, prompt initialization, and experimental configuration, while replacing only the underlying LLM. The complete FedEHR-Agents framework is compared with its PromptAvg variant under each setting.

Predictive performance generally improves with more capable LLM backbones for both methods (Appendix~Fig.~\ref{fig:llm_backbones}), indicating that stronger reasoning capabilities can benefit automated EHR modeling. More importantly, FedEHR-Agents consistently outperforms PromptAvg across all evaluated backbones, demonstrating that the advantage of evidence-guided experience aggregation is not specific to a particular LLM. The consistent performance gains suggest that EGEA can effectively integrate cross-hospital modeling experience under different reasoning backbones, while allowing the overall framework to benefit from advances in LLM capability.

\section{Discussion and Future Work}
\label{sec:discussion}

FedEHR-Agents advances toward federated autonomous clinical intelligence by preserving patient data locality. All patient-level EHR records and locally developed prediction models remain within their originating hospitals, while cross-hospital collaboration occurs via shareable experience representations (refined prompts and structured modeling evidence) and local historical memory. Although these representations exclude raw patient records, they may indirectly encode local data characteristics—such as feature importances, validation statistics, and optimization patterns—under strong inference or reconstruction attacks. Consequently, experience sharing should be viewed as a privacy-aware collaborative mechanism rather than a formal guarantee against privacy leakage.

To address these limitations and broaden the scope of autonomous healthcare AI, future research will explore several key directions:

\begin{itemize}
    \item \textbf{Privacy-Preserving Experience Communication:} Integrating formal privacy guarantees into experience-level updates using differential privacy for structured modeling evidence~\cite{brendan2018learning, andrew2021differentially}, secure aggregation techniques~\cite{bonawitz2017practical, shao2022dres}, trusted execution environments for server-side aggregation~\cite{kalapaaking2022blockchain}, and locally deployed LLMs to prevent external service exposure.
    \item \textbf{Multimodal and Adaptive Capabilities:} Extending the framework to handle multimodal clinical data (combining structured EHRs, medical imaging, and clinical text) and developing adaptive mechanisms for experience selection, aggregation, and continual evolution across heterogeneous institutions.
    \item \textbf{Long-Term Collaboration Assessment:} Conducting systematic, long-term evaluations of multi-agent collaboration and continual learning to assess sustained reliability and safety in real-world deployment.
\end{itemize}

\section{Acknowledgments}
Y.L. is supported by NOVA-FRQNT-NSERC grant (FRQ-NT 2023-NOVA-328677), Canada Research Chair (Tier 2) in Machine Learning for Genomics and Healthcare (CRC-2021-00547) and Natural Sciences and Engineering Research Council (NSERC) Discovery Grant (RGPIN-2016-05174).

\bibliographystyle{unsrt}
\bibliography{refs}

@String{Computing = "Computing" }

@String{Computer = "{IEEE} Computer" }

@article{johnson2016mimic,
  title={MIMIC-III, a freely accessible critical care database},
  author={Johnson, Alistair EW and Pollard, Tom J and Shen, Lu and Lehman, Li-wei H and Feng, Mengling and Ghassemi, Mohammad and Moody, Benjamin and Szolovits, Peter and Anthony Celi, Leo and Mark, Roger G},
  journal={Scientific data},
  volume={3},
  number={1},
  pages={1--9},
  year={2016},
  publisher={Nature Publishing Group}
}

@article{pollard2018eicu,
  title={The eICU Collaborative Research Database, a freely available multi-center database for critical care research},
  author={Pollard, Tom J and Johnson, Alistair EW and Raffa, Jesse D and Celi, Leo A and Mark, Roger G and Badawi, Omar},
  journal={Scientific data},
  volume={5},
  number={1},
  pages={180178},
  year={2018},
  publisher={Nature Publishing Group}
}

@article{shickel2017deep,
  title={Deep EHR: a survey of recent advances in deep learning techniques for electronic health record (EHR) analysis},
  author={Shickel, Benjamin and Tighe, Patrick James and Bihorac, Azra and Rashidi, Parisa},
  journal={IEEE journal of biomedical and health informatics},
  volume={22},
  number={5},
  pages={1589--1604},
  year={2017},
  publisher={IEEE}
}

@article{rajkomar2018scalable,
  title={Scalable and accurate deep learning with electronic health records},
  author={Rajkomar, Alvin and Oren, Eyal and Chen, Kai and Dai, Andrew M and Hajaj, Nissan and Hardt, Michaela and Liu, Peter J and Liu, Xiaobing and Marcus, Jake and Sun, Mimi and others},
  journal={NPJ digital medicine},
  volume={1},
  number={1},
  pages={18},
  year={2018},
  publisher={Nature Publishing Group UK London}
}

@article{kaissis2020secure,
  title={Secure, privacy-preserving and federated machine learning in medical imaging},
  author={Kaissis, Georgios A and Makowski, Marcus R and R{\"u}ckert, Daniel and Braren, Rickmer F},
  journal={Nature Machine Intelligence},
  volume={2},
  number={6},
  pages={305--311},
  year={2020},
  publisher={Nature Publishing Group UK London}
}

@article{rieke2020future,
  title={The future of digital health with federated learning},
  author={Rieke, Nicola and Hancox, Jonny and Li, Wenqi and Milletari, Fausto and Roth, Holger R and Albarqouni, Shadi and Bakas, Spyridon and Galtier, Mathieu N and Landman, Bennett A and Maier-Hein, Klaus and others},
  journal={NPJ digital medicine},
  volume={3},
  number={1},
  pages={119},
  year={2020},
  publisher={Nature Publishing Group UK London}
}

@inproceedings{mcmahan2017fedavg,
  title={Communication-efficient learning of deep networks from decentralized data},
  author={McMahan, Brendan and Moore, Eider and Ramage, Daniel and Hampson, Seth and y Arcas, Blaise Aguera},
  booktitle={Artificial intelligence and statistics},
  pages={1273--1282},
  year={2017},
  organization={Pmlr}
}

@article{li2020fedprox,
  title={Federated optimization in heterogeneous networks},
  author={Li, Tian and Sahu, Anit Kumar and Zaheer, Manzil and Sanjabi, Maziar and Talwalkar, Ameet and Smith, Virginia},
  journal={Proceedings of Machine learning and systems},
  volume={2},
  pages={429--450},
  year={2020}
}

@inproceedings{karimireddy2020scaffold,
  title={Scaffold: Stochastic controlled averaging for federated learning},
  author={Karimireddy, Sai Praneeth and Kale, Satyen and Mohri, Mehryar and Reddi, Sashank and Stich, Sebastian and Suresh, Ananda Theertha},
  booktitle={International conference on machine learning},
  pages={5132--5143},
  year={2020},
  organization={PMLR},
  publisher={},
  address={}
}

@article{wang2020fednova,
  title={Tackling the objective inconsistency problem in heterogeneous federated optimization},
  author={Wang, Jianyu and Liu, Qinghua and Liang, Hao and Joshi, Gauri and Poor, H Vincent},
  journal={Advances in neural information processing systems},
  volume={33},
  pages={7611--7623},
  year={2020}
}

@inproceedings{acar2021feddyn,
  title={Federated learning based on dynamic regularization},
  author={Durmus, Alp Emre and Yue, Zhao and Ramon, Matas and Matthew, Mattina and Paul, Whatmough and Venkatesh, Saligrama},
  booktitle={International conference on learning representations},
  year={2021},
}

@article{holzinger2019causability,
  title={Causability and explainability of artificial intelligence in medicine},
  author={Holzinger, Andreas and Langs, Georg and Denk, Helmut and Zatloukal, Kurt and M{\"u}ller, Heimo},
  journal={Wiley interdisciplinary reviews: data mining and knowledge discovery},
  volume={9},
  number={4},
  pages={e1312},
  year={2019},
  publisher={Wiley Online Library}
}

@article{topol2019high,
  title={High-performance medicine: the convergence of human and artificial intelligence},
  author={Topol, Eric J},
  journal={Nature medicine},
  volume={25},
  number={1},
  pages={44--56},
  year={2019},
  publisher={Nature Publishing Group US New York}
}

@inproceedings{
yao2023react,
title={ReAct: Synergizing Reasoning and Acting in Language Models},
author={Shunyu Yao and Jeffrey Zhao and Dian Yu and Nan Du and Izhak Shafran and Karthik R Narasimhan and Yuan Cao},
booktitle={The Eleventh International Conference on Learning Representations },
year={2023},
publisher={},
address={},
url={https://openreview.net/forum?id=WE_vluYUL-X},

}

@inproceedings{
shinn2023reflexion,
title={Reflexion: language agents with verbal reinforcement learning},
author={Noah Shinn and Federico Cassano and Ashwin Gopinath and Karthik R Narasimhan and Shunyu Yao},
booktitle={Thirty-seventh Conference on Neural Information Processing Systems},
year={2023},
url={https://openreview.net/forum?id=vAElhFcKW6}
}

@article{yuksekgonul2024textgrad,
  title={Textgrad: Automatic" differentiation" via text},
  author={Yuksekgonul, Mert and Bianchi, Federico and Boen, Joseph and Liu, Sheng and Huang, Zhi and Guestrin, Carlos and Zou, James},
  journal={arXiv preprint arXiv:2406.07496},
  year={2024}
}

@article{guo2024llmagent,
  title={Large language model based multi-agents: A survey of progress and challenges},
  author={Guo, Taicheng and Chen, Xiuying and Wang, Yaqi and Chang, Ruidi and Pei, Shichao and Chawla, Nitesh V and Wiest, Olaf and Zhang, Xiangliang},
  journal={arXiv preprint arXiv:2402.01680},
   volume={},
  number={},
  pages={},
  year={2024}
}

@inproceedings{bai2025unified,
  title={A unified solution to diverse heterogeneities in one-shot federated learning},
  author={Bai, Jun and Song, Yiliao and Wu, Di and Sajjanhar, Atul and Xiang, Yong and Zhou, Wei and Tao, Xiaohui and Li, Yan and Li, Yue},
  booktitle={Proceedings of the 31st ACM SIGKDD Conference on Knowledge Discovery and Data Mining V. 2},
  pages={71--82},
  year={2025},
   publisher={ACM},
  address={New York, NY, USA},
}

@article{ren2025healthcare,
  title={Healthcare agent: eliciting the power of large language models for medical consultation},
  author={Ren, Zhiyao and Zhan, Yibing and Yu, Baosheng and Ding, Liang and Xu, Pingbo and Tao, Dacheng},
  journal={npj Artificial Intelligence},
  volume={1},
  number={1},
  pages={24},
  year={2025},
  publisher={Nature Publishing Group UK London}
}

@article{ferber2026towards,
  title={Towards autonomous medical artificial intelligence agents},
  author={Ferber, Dyke and Hilgers, Lars and H{\"o}per, Christiane and Kinny-K{\"o}ster, Benedict and Eckardt, Jan-Niklas and Egger-Heidrich, Katharina and Bill, Marius and Schneider, Martin MK and Clusmann, Jan and Kadric, Lejla and others},
  journal={Nature},
  pages={1--10},
  year={2026},
  publisher={Nature Publishing Group}
}

@article{wang2025agent,
  title={Agent-based feature generation from clinical notes for outcome prediction},
  author={Wang, Jiayi and Vallon, Jacqueline Jil and Panjwani, Neil and Ling, Xi and Vij, Sushmita and Srinivas, Sandy and Leppert, John and Buyyounouski, Mark K and Bayati, Mohsen},
  journal={arXiv e-prints},
  pages={arXiv--2508},
  year={2025}
}

@article{zhu2025fedweight,
  title={FedWeight: mitigating covariate shift of federated learning on electronic health records data through patients re-weighting},
  author={Zhu, He and Bai, Jun and Li, Na and Li, Xiaoxiao and Liu, Dianbo and Buckeridge, David L and Li, Yue},
  journal={npj Digital Medicine},
  volume={8},
  number={1},
  pages={286},
  year={2025},
  publisher={Nature Publishing Group UK London}
}

@inproceedings{
trirat2025automlagent,
title={Auto{ML}-Agent: A Multi-Agent {LLM} Framework for Full-Pipeline Auto{ML}},
author={Patara Trirat and Wonyong Jeong and Sung Ju Hwang},
booktitle={Forty-second International Conference on Machine Learning},
year={2025},
url={https://openreview.net/forum?id=p1UBWkOvZm}
}

@article{hunik2025diagnostic,
  title={Diagnostic prediction models for primary care, based on AI and electronic health records: systematic review},
  author={Hunik, Liesbeth and Chaabouni, Asma and van Laarhoven, Twan and Hartman, Tim C Olde and Leijenaar, Ralph TH and Cals, Jochen WL and Uijen, Annemarie A and Schers, Henk J},
  journal={JMIR Medical Informatics},
  volume={13},
  number={1},
  pages={e62862},
  year={2025},
  publisher={JMIR Publications Inc., Toronto, Canada}
}

@article{teo2025generative,
  title={Generative artificial intelligence in medicine},
  author={Teo, Zhen Ling and Thirunavukarasu, Arun James and Elangovan, Kabilan and Cheng, Haoran and Moova, Prasanth and Soetikno, Brian and Nielsen, Christopher and Pollreisz, Andreas and Ting, Darren Shu Jeng and Morris, Robert JT and others},
  journal={Nature medicine},
  volume={31},
  number={10},
  pages={3270--3282},
  year={2025},
  publisher={Nature Publishing Group US New York}
}

@article{martinek2026agentomics,
  title={Agentomics: an agentic system that autonomously develops novel state-of-the-art solutions for biomedical machine learning tasks},
  author={Martinek, Vlastimil and Gariboldi, Andrea and Tzimotoudis, Dimosthenis and Galea, Mark and Zacharopoulou, Elissavet and Escudero, Aitor Alberdi and Blake, Edward and {\v{C}}ech{\'a}k, David and Cassar, Luke and Balestrucci, Alessandro and others},
  journal={Bioinformatics},
  volume={42},
  number={Supplement\_1},
  pages={btag250},
  year={2026},
  publisher={Oxford University Press}
}

@inproceedings{
brendan2018learning,
title={Learning Differentially Private Recurrent Language Models},
author={H. Brendan McMahan and Daniel Ramage and Kunal Talwar and Li Zhang},
booktitle={International Conference on Learning Representations},
year={2018},
url={https://openreview.net/forum?id=BJ0hF1Z0b},
}

@article{andrew2021differentially,
  title={Differentially private learning with adaptive clipping},
  author={Andrew, Galen and Thakkar, Om and McMahan, Brendan and Ramaswamy, Swaroop},
  journal={Advances in neural information processing systems},
  volume={34},
  pages={17455--17466},
  year={2021}
}

@inproceedings{bonawitz2017practical,
  title={Practical secure aggregation for privacy-preserving machine learning},
  author={Bonawitz, Keith and Ivanov, Vladimir and Kreuter, Ben and Marcedone, Antonio and McMahan, H Brendan and Patel, Sarvar and Ramage, Daniel and Segal, Aaron and Seth, Karn},
  booktitle={proceedings of the 2017 ACM SIGSAC Conference on Computer and Communications Security},
  pages={1175--1191},
  year={2017},
 publisher = {Association for Computing Machinery},
 address = {New York, NY, USA},
 isbn = {9781450349468},
}

@article{shao2022dres,
  title={Dres-fl: Dropout-resilient secure federated learning for non-iid clients via secret data sharing},
  author={Shao, Jiawei and Sun, Yuchang and Li, Songze and Zhang, Jun},
  journal={Advances in Neural Information Processing Systems},
  volume={35},
  pages={10533--10545},
  year={2022}
}

@article{kalapaaking2022blockchain,
  title={Blockchain-based federated learning with secure aggregation in trusted execution environment for internet-of-things},
  author={Kalapaaking, Aditya Pribadi and Khalil, Ibrahim and Rahman, Mohammad Saidur and Atiquzzaman, Mohammed and Yi, Xun and Almashor, Mahathir},
  journal={IEEE Transactions on Industrial Informatics},
  volume={19},
  number={2},
  pages={1703--1714},
  year={2022},
  publisher={IEEE}
}

@article{wang2026doctoragents,
  title={DoctorAgents: an agentic framework to iteratively refine AutoML pipeline for small clinical temporal data},
  author={Wang, Ruilin and Wang, Bo-Hong and Kourbatski, Elizabeth and Bai, Jun and Chen, Hegang and Song, Ziyang and Boire, Gilles and Hudson, Marie and Li, Yue},
  journal={arXiv preprint arXiv:2608.05375},
  year={2026}
}

\clearpage

\appendix

\counterwithin{figure}{section}
\counterwithin{table}{section}

\renewcommand{\thefigure}{\thesection\arabic{figure}}
\renewcommand{\thetable}{\thesection\arabic{table}}

\onecolumn

\begin{center}
  {\LARGE\bfseries Appendix}
\end{center}

\section{Overall Federated Optimization Procedure}
\label{appendix:algorithm}

\begin{algorithm}[!h]
\small
\caption{FedEHR-Agents}
\label{alg:fedehr_agents}
\begin{algorithmic}[1]

\Require Hospitals $\{H_k\}_{k=1}^{K}$ with private EHR datasets
$\{\mathcal{D}_k\}_{k=1}^{K}$, communication rounds $R$
\Ensure Refined clinical modeling experience and local prediction models

\State Initialize DP prompts
$\{\mathcal{P}_{k,\mathrm{DP}}^{(0)}\}_{k=1}^{K}$
\State Initialize MD prompts
$\{\mathcal{P}_{k,\mathrm{MD}}^{(0)}\}_{k=1}^{K}$
and historical memories
$\{\mathcal{M}_{k}^{(0)}\}_{k=1}^{K}$

\For{$r=0,\ldots,R-1$}

    \ForAll{hospitals $H_k$ \textbf{in parallel}}

        \State Run the DP module
        \[
        \mathbf{Z}_{k}^{(r)}
        \leftarrow
        g_{\mathrm{DP}}
        \left(
        \mathcal{D}_k,
        \mathcal{P}_{k,\mathrm{DP}}^{(r)}
        \right)
        \]

        \State Run the MD module
        \[
        f_k^{(r)}
        \leftarrow
        g_{\mathrm{MD}}
        \left(
        \mathbf{Z}_{k}^{(r)},
        \mathbf{y}_k,
        \mathcal{P}_{k,\mathrm{MD}}^{(r)}
        \right)
        \]

        \State Collect structured evidence
        $\mathcal{S}_{k,\mathrm{DP}}^{(r)}$ and
        $\mathcal{S}_{k,\mathrm{MD}}^{(r)}$

        \State Evaluate DP and MD outputs to obtain
        $\mathcal{F}_{k,\mathrm{DP}}^{(r)}$ and
        $\mathcal{F}_{k,\mathrm{MD}}^{(r)}$

        \State Refine the DP prompt via TextGrad
        \[
        \widetilde{\mathcal{P}}_{k,\mathrm{DP}}^{(r)}
        \leftarrow
        \Omega_{\mathrm{DP}}
        \left(
        \mathcal{P}_{k,\mathrm{DP}}^{(r)},
        \mathcal{F}_{k,\mathrm{DP}}^{(r)},
        \mathcal{M}_{k}^{(r)}
        \right)
        \]

        \State Refine the MD prompt via TextGrad
        \[
        \widetilde{\mathcal{P}}_{k,\mathrm{MD}}^{(r)}
        \leftarrow
        \Omega_{\mathrm{MD}}
        \left(
        \mathcal{P}_{k,\mathrm{MD}}^{(r)},
        \mathcal{F}_{k,\mathrm{MD}}^{(r)},
        \mathcal{M}_{k}^{(r)}
        \right)
        \]

        \State Update local historical memory $\mathcal{M}_{k}^{(r)}$

        \State Build the local DP experience representation
        \[
        \mathcal{R}_{k,\mathrm{DP}}^{(r)}
        \leftarrow
        \left(
        \widetilde{\mathcal{P}}_{k,\mathrm{DP}}^{(r)},
        \mathcal{S}_{k,\mathrm{DP}}^{(r)}
        \right)
        \]
        
        \State Build the local MD experience representation
        \[
        \mathcal{R}_{k,\mathrm{MD}}^{(r)}
        \leftarrow
        \left(
        \widetilde{\mathcal{P}}_{k,\mathrm{MD}}^{(r)},
        \mathcal{S}_{k,\mathrm{MD}}^{(r)}
        \right)
        \]
        
        \State Upload
        $\mathcal{R}_{k,\mathrm{DP}}^{(r)}$
        and
        $\mathcal{R}_{k,\mathrm{MD}}^{(r)}$

    \EndFor

    \State Aggregate DP experience
    \[
    \mathcal{R}_{G,\mathrm{DP}}^{(r)}
    \leftarrow
    \Psi_{\mathrm{DP}}
    \left(
    \mathcal{R}_{1,\mathrm{DP}}^{(r)},
    \ldots,
    \mathcal{R}_{K,\mathrm{DP}}^{(r)}
    \right)
    \]

    \State Aggregate MD experience
    \[
    \mathcal{R}_{G,\mathrm{MD}}^{(r)}
    \leftarrow
    \Psi_{\mathrm{MD}}
    \left(
    \mathcal{R}_{1,\mathrm{MD}}^{(r)},
    \ldots,
    \mathcal{R}_{K,\mathrm{MD}}^{(r)}
    \right)
    \]

    \State Generate global DP and MD meta-prompts
    \[
    \mathcal{P}_{G,m}^{(r)}
    \leftarrow
    \Gamma_m
    \left(
    \mathcal{R}_{G,m}^{(r)}
    \right),
    \quad
    m\in\{\mathrm{DP},\mathrm{MD}\}
    \]

    \ForAll{hospitals $H_k$ \textbf{in parallel}}

        \State Integrate the global meta-prompts
        \[
        \mathcal{P}_{k,m}^{(r+1)}
        \leftarrow
        \Phi_m
        \left(
        \widetilde{\mathcal{P}}_{k,m}^{(r)},
        \mathcal{P}_{G,m}^{(r)}
        \right),
        \quad
        m\in\{\mathrm{DP},\mathrm{MD}\}
        \]

    \EndFor

\EndFor

\State \Return
$\{
\mathcal{P}_{k,\mathrm{DP}}^{(R)},
\mathcal{P}_{k,\mathrm{MD}}^{(R)},
\mathcal{M}_{k}^{(R)},
f_k^{(R-1)}
\}_{k=1}^{K}$

\end{algorithmic}
\end{algorithm}

\section{Additional Experimental Results}
\label{appendix:additional_results}

\subsection{Scalability Across Additional Clinical Tasks}
\label{appendix:scalability_tasks}

To further examine whether the scalability observed on Mortality-48h generalizes to other clinical prediction tasks, we extend the analysis to ARF-4h, LOS$>$7d, and Sepsis. Following the same experimental setting, the number of participating hospitals is varied from 3 to 20, and the average AUPRC across participating hospitals is reported. FedEHR-Agents is compared with its PromptAvg variant under each federation scale (Table~\ref{tab:additional_scalability}).

Both methods generally benefit from increasing the number of participating hospitals across all three tasks. More importantly, FedEHR-Agents consistently outperforms PromptAvg at every federation scale (Table~\ref{tab:additional_scalability}). As the number of hospitals increases from 3 to 20, the AUPRC of FedEHR-Agents improves from 0.194 to 0.236 on ARF-4h, from 0.532 to 0.553 on LOS$>$7d, and from 0.316 to 0.341 on Sepsis. At 20 hospitals, FedEHR-Agents exceeds PromptAvg by 0.022, 0.020, and 0.023 AUPRC on the three tasks, respectively. These results are consistent with the Mortality-48h findings in the main text, indicating that evidence-guided experience aggregation remains effective as the federation expands across different clinical prediction tasks.

\begin{table*}[t]
\centering
\caption{
Scalability of FedEHR-Agents across additional clinical prediction tasks.
Average AUPRC is reported under different numbers of participating hospitals. FedEHR-Agents (PromptAvg) is abbreviated as PromptAvg.
}
\label{tab:additional_scalability}
\begin{tabular}{c cc cc cc}
\toprule
\multirow{2}{*}{\textbf{\# Hospitals}}
& \multicolumn{2}{c}{\textbf{ARF-4h}}
& \multicolumn{2}{c}{\textbf{LOS$>$7d}}
& \multicolumn{2}{c}{\textbf{Sepsis}} \\
\cmidrule(lr){2-3}
\cmidrule(lr){4-5}
\cmidrule(lr){6-7}
& PromptAvg & FedEHR-Agents
& PromptAvg & FedEHR-Agents
& PromptAvg & FedEHR-Agents \\
\midrule
3  & 0.183 ± 0.006 &  0.194 ± 0.004 & 0.506 ± 0.012 & 0.532 ± 0.010  & 0.286 ± 0.011 & 0.316 ± 0.009 \\
5  & 0.197 ± 0.004 & 0.225 ± 0.003 & 0.513 ± 0.011 & 0.538 ± 0.009 & 0.302 ± 0.011 & 0.322 ± 0.011 \\
10 & 0.202 ± 0.006 & 0.229 ± 0.005 & 0.527 ± 0.010 & 0.544 ± 0.010 & 0.311 ± 0.010 & 0.328 ± 0.010 \\
15 & 0.208 ± 0.005 & 0.231 ± 0.005 & 0.529 ± 0.013 & 0.546 ± 0.008 & 0.317 ± 0.007 & 0.335 ± 0.009 \\
20 & 0.214 ± 0.006 & 0.236 ± 0.004 & 0.533 ± 0.011 & 0.553 ± 0.011 & 0.318 ± 0.009 & 0.341 ± 0.010 \\
\bottomrule
\end{tabular}
\end{table*}

\subsection{Ablation Across Additional Clinical Tasks}
\label{appendix:ablation_tasks}

Removing any individual component consistently degrades the performance of FedEHR-Agents across ARF-4h, LOS$>$7d, and Sepsis (Table~\ref{tab:additional_ablation}). The DP Module contributes most strongly to ARF-4h, where its removal decreases the performance from 0.194 to 0.176. In comparison, EGEA has the largest impact on LOS$>$7d and Sepsis, with performance dropping from 0.532 to 0.506 and from 0.316 to 0.286, respectively. Removing the Evaluator also leads to consistent performance degradation across all three tasks, while Memory provides smaller but stable improvements. Overall, the complete FedEHR-Agents achieves the best performance on all evaluated tasks, demonstrating that these components provide complementary contributions across different clinical prediction settings.

\begin{table*}[!h]
\centering
\caption{
Ablation results of FedEHR-Agents across additional clinical prediction tasks.
}
\label{tab:additional_ablation}
\begin{tabular}{cccc|ccc}
\toprule
\textbf{Memory}
& \textbf{Evaluator}
& \textbf{DP Module}
& \textbf{EGEA}
& \textbf{ARF-4h}
& \textbf{LOS$>$7d}
& \textbf{Sepsis} \\
\midrule
$\times$  & \checkmark & \checkmark & \checkmark
& 0.187 ± 0.004 & 0.521 ± 0.012 & 0.303 ± 0.010 \\

\checkmark & $\times$  & \checkmark & \checkmark
& 0.185 ± 0.005 & 0.515 ± 0.010 & 0.298 ± 0.009 \\

\checkmark & \checkmark & $\times$  & \checkmark
& 0.176 ± 0.005 & 0.510 ± 0.011 & 0.294 ± 0.008 \\

\checkmark & \checkmark & \checkmark & $\times$
& 0.183 ± 0.006 & 0.506 ± 0.010  & 0.286 ± 0.011 \\

\midrule
\checkmark & \checkmark & \checkmark & \checkmark
& \textbf{0.194 ± 0.004} & \textbf{0.532 ± 0.010 } & \textbf{0.316 ± 0.009} \\
\bottomrule
\end{tabular}
\end{table*}

\subsection{Robustness to Different LLM Backbones}
\label{appendix:llm}

\begin{figure}[!h]
    \centering
    \includegraphics[width=0.5\linewidth]{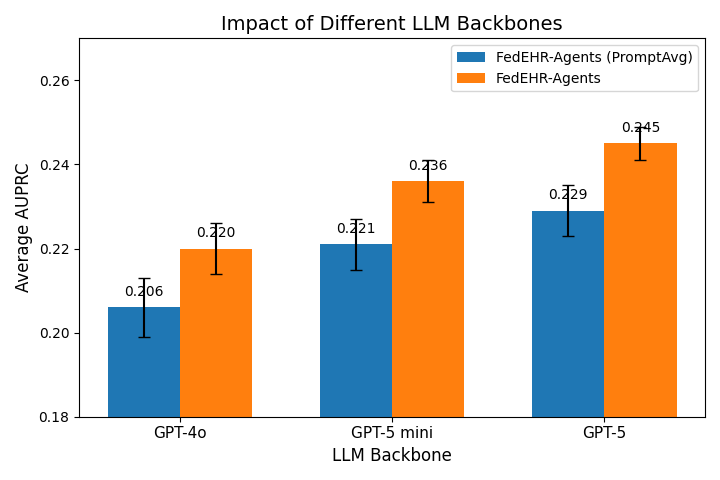}
    \caption{
    Comparison of different LLM backbones for federated clinical agent optimization. Results are averaged across the three participating hospitals.
    }
    \Description{.}
    \label{fig:llm_backbones}
\end{figure}

We further evaluate FedEHR-Agents with GPT-4o, GPT-5 mini, and GPT-5 while keeping all other experimental settings unchanged. FedEHR-Agents consistently outperforms its variant, FedEHR-Agents (PromptAvg), across all evaluated LLM backbones, demonstrating its robustness to the choice of the underlying LLM (Fig.~\ref{fig:llm_backbones}).

\section{Dataset and Feature Configuration}
\label{appendix:data_configuration}

All experiments are conducted on the eICU Collaborative Research Database, where each participating hospital is treated as an independent federated client. For each hospital, we extract information from the major clinical sources used in our experiments, including demographics, intake/output records, laboratory measurements, medications, respiratory-care records, and periodic and aperiodic vital signs. Task-specific labels are separately constructed according to the corresponding clinical prediction objective.

The original eICU tables exhibit substantially different schemas and data organizations. To provide a consistent interface for autonomous data preprocessing, we first transform each clinical source into a unified longitudinal event representation,
$
(\texttt{ID},\ \texttt{t},\ 
    \texttt{variable\_name},\ 
    \texttt{variable\_value}),
    \label{eq:standardized_ehr_format}
$
where \texttt{ID} identifies the patient or ICU stay, \texttt{t} denotes the observation time, and \texttt{variable\_name} and 
\texttt{variable\_value} specify the corresponding clinical observation. This schema-level standardization converts heterogeneous EHR sources into a common event-based format while preserving their temporal information. Importantly, this step does not impose a predefined feature engineering strategy. Instead, it decouples subsequent agent-based feature construction 
from table-specific database schemas, allowing the DP module to apply a common processing interface for temporal alignment, aggregation, missing-value handling, and feature selection across heterogeneous clinical sources.

At communication round $r$, the DP module independently processes each standardized clinical source $s$ and constructs a task-specific temporal feature representation,
$
    \mathbf{Z}_{k,s}^{(r)}
    \in
    \mathbb{R}^{N_k \times T \times D_{k,s}^{(r)}},
    \label{eq:source_temporal_features}
$
where $N_k$ denotes the number of patient samples at hospital $k$, $T$ is the number of temporal steps defined by the prediction setting, and $D_{k,s}^{(r)}$ denotes the number of features generated or retained from clinical source $s$. The source-specific representations are then temporally 
aligned and concatenated along the feature dimension,
\begin{equation}
    \mathbf{Z}_{k}^{(r)}
    =
    \operatorname{Concat}
    \left(
        \mathbf{Z}_{k,1}^{(r)},
        \ldots,
        \mathbf{Z}_{k,S}^{(r)}
    \right)
    \in
    \mathbb{R}^{N_k \times T \times D_k^{(r)}},
    \label{eq:fused_temporal_features}
\end{equation}
where
$D_k^{(r)}=\sum_{s=1}^{S}D_{k,s}^{(r)}$
denotes the resulting feature dimension. For the Mortality-48h task, all temporal observations are organized into $T=48$ hourly steps.

At the first communication round, the DP module constructs an initial task-specific feature space independently for each hospital. Table~\ref{tab:initial_feature_configuration} summarizes the
corresponding sample sizes and source-level feature composition for three
representative hospitals. Hospitals 73, 264, and 420 contain 2,663, 3,006, and 2,636 patient samples, respectively, while their initial feature spaces contain 112, 98, and 108 features. The resulting feature dimensions are therefore allowed to differ across hospitals rather than being manually aligned to a common feature space.

\begin{table*}[!h]
\centering
\caption{
Initial feature composition of three representative hospitals on the
Mortality-48h task. $N_k$ denotes the number of patient samples, and the
remaining columns report the numbers of features initially retained from
different clinical sources at the first communication round.
}
\label{tab:initial_feature_configuration}
\small
\setlength{\tabcolsep}{4pt}
\begin{tabular}{lccccccccc}
\toprule
\textbf{Hospital}
& $\mathbf{N_k}$
& \shortstack{\textbf{Demographic}\\\textbf{Features}}
& \shortstack{\textbf{Intake/Output}\\\textbf{Features}}
& \shortstack{\textbf{Laboratory}\\\textbf{Features}}
& \shortstack{\textbf{Medication}\\\textbf{Features}}
& \shortstack{\textbf{Respiratory Care}\\\textbf{Features}}
& \shortstack{\textbf{Periodic Vital}\\\textbf{Sign Features}}
& \shortstack{\textbf{Aperiodic Vital}\\\textbf{Sign Features}}
& \textbf{Total} \\
\midrule
Hospital 73
& 2,663
& 10
& 4
& 38
& 18
& 16
& 16
& 10
& 112 \\

Hospital 264
& 3,006
& 9
& 4
& 33
& 15
& 14
& 14
& 9
& 98 \\

Hospital 420
& 2,636
& 10
& 4
& 37
& 17
& 15
& 15
& 10
& 108 \\
\bottomrule
\end{tabular}
\end{table*}

The source-level feature composition varies across hospitals, reflecting
differences in locally available clinical variables, measurement frequency, missingness patterns, and recording practices. Laboratory measurements typically contribute a relatively large portion of the initial feature space, while medications, respiratory-care variables, vital signs, and demographic information provide complementary clinical information. Rather than eliminating this heterogeneity through manually harmonized feature engineering, FedEHR-Agents preserves hospital-specific feature spaces and allows the DP module to adapt subsequent feature construction and selection to each local EHR environment. This setting more closely reflects real multi-hospital deployment, where both the availability and informativeness of clinical variables may differ substantially across institutions.

\subsection{Clinical Feature Definitions for Mortality-48h}
\label{appendix:feature_definitions}

To facilitate interpretation of the SHAP analysis in Fig.~\ref{fig:shap_analysis}, Table~\ref{tab:mortality_feature_definitions} summarizes the eICU mappings and clinical meanings of the features reported for the Mortality-48h task. The feature names displayed in Fig.~\ref{fig:shap_analysis} are standardized names used after preprocessing and therefore do not necessarily preserve the exact naming convention of the original eICU variables. Most features can be directly mapped to laboratory, demographic, or vital-sign variables in eICU, while a small number are
constructed from the corresponding source variables during preprocessing.

\begin{table*}[t]
\centering
\caption{
Definitions and eICU mappings of the clinical features reported in the
Mortality-48h SHAP analysis.
}
\label{tab:mortality_feature_definitions}
\footnotesize
\setlength{\tabcolsep}{4pt}
\renewcommand{\arraystretch}{1.08}
\begin{tabular}{
p{0.13\textwidth}
p{0.24\textwidth}
p{0.16\textwidth}
p{0.40\textwidth}}
\toprule
\textbf{Feature Name}
& \textbf{eICU Mapping}
& \textbf{Table Source}
& \textbf{Description} \\
\midrule

\texttt{creatinine}
& \texttt{lab: creatinine}
& Laboratory
& Serum creatinine reflecting renal function and potential kidney impairment. \\

\texttt{bun}
& \texttt{lab: BUN}
& Laboratory
& Blood urea nitrogen reflecting renal function and accumulation of nitrogenous waste products. \\

\texttt{lactate}
& \texttt{lab: lactate}
& Laboratory
& Blood lactate reflecting tissue hypoperfusion and metabolic stress. \\

\texttt{glucose}
& \texttt{lab: glucose / bedside glucose}
& Laboratory
& Blood glucose concentration reflecting glycemic and metabolic status. \\

\texttt{wbc}
& \texttt{lab: WBC x 1000}
& Laboratory
& White blood cell count reflecting inflammatory and infectious responses. \\

\texttt{age}
& \texttt{patient.age}
& Demographics
& Patient age at ICU admission, representing baseline demographic risk. \\

\texttt{anion\_gap}
& \texttt{lab: anion gap}
& Laboratory
& Anion gap reflecting acid--base imbalance and potential metabolic acidosis. \\

\texttt{heartrate}
& \texttt{vitalPeriodic.heartRate}
& Periodic Vital Signs
& Heart rate reflecting cardiovascular activity and hemodynamic status. \\

\texttt{potassium}
& \texttt{lab: potassium}
& Laboratory
& Serum potassium concentration reflecting electrolyte balance and cardiac-related physiological status. \\

\texttt{respiration}
& \texttt{vitalPeriodic.respiration}
& Periodic Vital Signs
& Respiratory rate reflecting breathing frequency and overall respiratory status. \\

\texttt{bicarbonate}
& \texttt{lab: bicarbonate}
& Laboratory
& Serum bicarbonate reflecting metabolic and acid--base status. \\

\texttt{platelets}
& \texttt{lab: platelets x 1000}
& Laboratory
& Platelet count reflecting hematological and coagulation-related status. \\

\texttt{sao2}
& \texttt{vitalPeriodic.saO2}
& Periodic Vital Signs
& Oxygen saturation reflecting systemic oxygenation status. \\

\texttt{hemoglobin}
& \texttt{lab: Hgb}
& Laboratory
& Hemoglobin concentration reflecting blood oxygen-carrying capacity and hematological status. \\

\texttt{sodium}
& \texttt{lab: sodium}
& Laboratory
& Serum sodium concentration reflecting electrolyte and fluid balance. \\

\texttt{vasopressor}
& \texttt{medication.drugname} (derived)
& Medication
& Derived indicator of vasopressor administration, reflecting the requirement for pharmacological hemodynamic support. \\

\texttt{vent\_start}
& \texttt{respiratoryCare.ventStartOffset} (derived)
& Respiratory Care
& Derived indicator of ventilation initiation, reflecting the requirement for mechanical respiratory support. \\

\texttt{systemicmean}
& \texttt{vitalPeriodic.systemicMean}
& Periodic Vital Signs
& Mean systemic arterial pressure reflecting overall hemodynamic and tissue-perfusion status. \\

\bottomrule
\end{tabular}
\end{table*}

The reported features span multiple clinically relevant physiological dimensions, including renal function, metabolic and acid--base status, inflammation, electrolyte balance, cardiovascular and respiratory status, hematological measurements, and organ-support interventions. These definitions
provide the clinical context for interpreting the hospital-level SHAP
distributions reported in Fig.~\ref{fig:shap_analysis}.

\section{Prompt Templates and Structured Outputs}
\label{appendix:prompts}

his section provides the prompt templates used by the major functional components of FedEHR-Agents. Rather than relying on a single generic prompt, the framework adopts module-specific instructions for data preprocessing, model development, local evaluation and refinement, and server-side
experience aggregation. Each template specifies the corresponding role, available inputs, execution constraints, and expected outputs, while task- and hospital-specific information is dynamically inserted during execution.

For readability, the prompts below are presented in a consolidated form that preserves the instructions used in our implementation while removing repetitive engineering details. We additionally provide representative structured outputs to illustrate how execution results, evaluator feedback,
and modeling evidence are organized for subsequent local prompt refinement and federated experience aggregation. All information communicated outside the local execution environment is restricted to modeling-level information and does not contain patient-level EHR records.

\subsection{Data Preprocessing Prompts}
\label{appendix:dp_prompts}

The DP Module employs a two-level prompting strategy. A general system prompt specifies the common preprocessing objective, temporal representation, patient alignment, leakage constraints, and output interface. A source-specific instruction is then appended according to the clinical table being processed. This separation allows the same DP Module to handle heterogeneous EHR sources while adapting feature construction to their distinct temporal and semantic characteristics.

In our implementation, each longitudinal EHR source is represented in a standardized long format containing \texttt{ID}, \texttt{t}, \texttt{variable\_name}, and \texttt{variable\_value}. The DP Module converts each source independently into a temporal representation and subsequently supports their alignment and integration for downstream model development.

\paragraph{General DP System Prompt.}
The following prompt defines the common preprocessing requirements used across EHR sources, while task-, hospital-, and round-specific information is dynamically inserted through the input interface.

\begin{promptbox}{System Prompt for Data Preprocessing}

You are a clinical temporal data preprocessing code generator.

Your task is to generate one complete and executable Python program that transforms a longitudinal EHR source into a temporally structured representation for the specified clinical prediction task.

\textbf{TASK OBJECTIVE}

Construct an informative and leakage-safe temporal representation from the clinical observations available within the specified observation window.

The clinical prediction setting is defined by:

\begin{promptitemize}
    \item \texttt{Task: \{TASK\_NAME\}}
    \item \texttt{Target Outcome: \{TARGET\_OUTCOME\}}
    \item \texttt{Observation Window: \{OBSERVATION\_WINDOW\}}
    \item \texttt{Temporal Resolution: \{TEMPORAL\_RESOLUTION\}}
\end{promptitemize}

The target outcome describes the prediction objective only. Do not use patient-level labels for feature construction or selection.

\textbf{DYNAMIC INPUTS}

You will be provided with:

\begin{promptitemize}
    \item \texttt{Source Name: \{TABLE\_NAME\}}
    \item \texttt{Source Data Path: \{TABLE\_PATH\}}
    \item \texttt{Patient Cohort Path: \{COHORT\_PATH\}}
    \item \texttt{Input Schema: \{TABLE\_SCHEMA\}}
    \item \texttt{Output Tensor Path: \{TENSOR\_OUTPUT\_PATH\}}
    \item \texttt{Flattened Output Path: \{FLAT\_OUTPUT\_PATH\}}
    \item \texttt{Feature Name Path: \{FEATURE\_NAME\_PATH\}}
    \item \texttt{Global Meta-Prompt:
    \{GLOBAL\_DP\_META\_PROMPT\}}
    \item \texttt{Source-Specific Instruction:
    \{SOURCE\_SPECIFIC\_INSTRUCTION\}}
\end{promptitemize}

The global meta-prompt contains transferable preprocessing guidance obtained from previous cross-hospital collaboration. Use it as high-level guidance, but adapt the final preprocessing strategy to the characteristics of the current local EHR source.

The source-specific instruction defines the clinical and temporal characteristics of the current source and should guide the corresponding feature construction and aggregation strategy.

\textbf{INPUT FORMAT}

The standardized longitudinal EHR source follows:

\begin{center}
\texttt{ID, t, variable\_name, variable\_value}
\end{center}

where:

\begin{promptitemize}
    \item \texttt{ID} identifies the patient or ICU stay;
    \item \texttt{t} denotes observation time in minutes;
    \item \texttt{variable\_name} specifies the clinical variable;
    \item \texttt{variable\_value} stores the corresponding observation.
\end{promptitemize}

Multiple observations may occur for the same patient, variable, and temporal interval.

\textbf{ROLE BOUNDARY}

\begin{promptitemize}
    \item Focus only on temporal data preprocessing and feature construction.
    \item Do not perform predictive model development or model selection.
    \item Do not use patient-level target labels as input features or for feature construction.
    \item Preserve the clinical meaning and temporal ordering of the source observations.
\end{promptitemize}

\textbf{PREPROCESSING REQUIREMENTS}

\begin{promptenum}

\item \textbf{Temporal construction}

Convert the observation timestamps into the temporal resolution specified by \texttt{\{TEMPORAL\_RESOLUTION\}} and retain only observations within \texttt{\{OBSERVATION\_WINDOW\}}.

For hourly temporal representations, use:

\begin{center}
\texttt{hour\_index = t // 60}.
\end{center}

Do not use observations outside the permitted prediction window.

\item \textbf{Patient consistency}

Use \texttt{\{COHORT\_PATH\}} only to determine the complete patient cohort
and its ordering.

\begin{promptitemize}
    \item Preserve the exact patient ordering defined by the cohort.
    \item Ensure that all patients are represented in the output.
    \item If a patient has no records in the current source, use an appropriate empty or default representation.
    \item Do not use target-label values contained in the cohort file for preprocessing or feature construction.
\end{promptitemize}

\item \textbf{Source-aware feature construction}

Construct a compact and interpretable representation appropriate to the current EHR source.

\begin{promptitemize}
    \item Use \texttt{variable\_name} to identify the available clinical variables.
    \item Apply the provided source-specific instruction when determining aggregation, missing-value handling, and feature construction.
    \item Preserve clinically meaningful temporal information when supported by the available observations.
    \item Avoid excessive, unstable, or highly redundant feature generation.
\end{promptitemize}

\item \textbf{Missing and irregular observations}

\begin{promptitemize}
    \item Correctly handle repeated observations within the same temporal interval.
    \item Apply source-appropriate aggregation when multiple observations are present.
    \item Handle missing or irregular observations using stable strategies appropriate to the current source.
    \item Avoid undefined or numerically unstable output values.
\end{promptitemize}

\item \textbf{Data-leakage constraints}

\begin{promptitemize}
    \item Do not use patient-level prediction labels for feature construction or selection.
    \item Do not use future observations beyond the specified observation window.
    \item Do not construct features that directly or indirectly encode the
    target outcome.
\end{promptitemize}

\item \textbf{Output consistency}

Generate a temporal representation with shape

\[
(N,T,D_{\mathrm{source}}),
\]

where $N$ is the number of patients, $T$ is the number of temporal intervals, and $D_{\mathrm{source}}$ is the number of feature channels constructed from the current EHR source.

Save:

\begin{promptitemize}
    \item the temporal tensor to
    \texttt{\{TENSOR\_OUTPUT\_PATH\}};
    \item the flattened representation to
    \texttt{\{FLAT\_OUTPUT\_PATH\}};
    \item the ordered feature-name list to
    \texttt{\{FEATURE\_NAME\_PATH\}}.
\end{promptitemize}

The flattened representation must use time-major ordering: all feature channels at time step $t=0$ appear first, followed by all channels at $t=1$, and so forth.

The number and ordering of feature names must exactly match the flattened tensor representation.

\end{promptenum}

Do not perform feature scaling or normalization unless explicitly required by the source-specific instruction.

\textbf{OUTPUT CONSTRAINT}

Return only one executable Python program. Do not include explanations, Markdown text, or any content outside the generated code.

\end{promptbox}

\paragraph{Source-Specific Instructions.}

For each EHR source, a source-specific instruction is dynamically inserted into the general DP prompt through \texttt{\{SOURCE\_SPECIFIC\_INSTRUCTION\}}. It describes the source semantics, sampling characteristics, and appropriate feature-construction strategy.

\begin{promptbox}{Source-Specific Data Preprocessing Instruction}

\textbf{EHR SOURCE}

\texttt{\{TABLE\_NAME\}}

\textbf{DATA CHARACTERISTICS}

\texttt{\{TABLE\_CHARACTERISTICS\}}

\textbf{AVAILABLE VARIABLES}

\texttt{\{VARIABLE\_DESCRIPTION\}}

\textbf{SOURCE-SPECIFIC FEATURE CONSTRUCTION}

\texttt{\{FEATURE\_CONSTRUCTION\_RULES\}}

\textbf{REPEATED AND MISSING OBSERVATIONS}

\texttt{\{AGGREGATION\_AND\_MISSINGNESS\_RULES\}}

\textbf{FEATURE NAMING}

Use deterministic and interpretable feature names that preserve the underlying clinical variable and, when applicable, the transformation type and temporal index.

Ensure that the feature-name ordering exactly matches the flattened temporal representation.

\end{promptbox}

The source-specific instructions depend on the temporal and semantic characteristics of each clinical table. Table~\ref{tab:dp_source_prompts} summarizes the primary strategies used for the longitudinal EHR sources in our implementation.

\begin{table*}[t]
\centering
\caption{
Source-specific preprocessing strategies used by the DP Module.
}
\label{tab:dp_source_prompts}
\setlength{\tabcolsep}{5pt}
\begin{tabular}{
p{0.18\textwidth}
p{0.27\textwidth}
p{0.48\textwidth}}
\toprule
\textbf{EHR Source}
& \textbf{Data Characteristics}
& \textbf{Primary Feature Strategy} \\
\midrule

Periodic Vital Signs
& Dense and relatively high-frequency physiological measurements
& Representative within-interval values together with compact
temporal-change or variability features when sufficiently supported. \\

Laboratory
& Sparse and irregular laboratory observations with high missingness
& Representative measurements together with selected temporal-change
or variability features while limiting redundant feature expansion. \\

Medication
& Sparse event-based medication administration records
& Binary indicators representing whether each medication category is
administered within the corresponding temporal interval. \\

Intake/Output
& Repeated fluid intake, output, and balance measurements
& Temporal aggregation of intake, output, and net fluid-balance variables
within each interval. \\

Respiratory Care
& Sparse but clinically important respiratory-intervention events
& Binary indicators representing the occurrence of respiratory-support
events within each temporal interval. \\

Aperiodic Vital Signs
& Sparse and intermittently measured physiological observations
& Representative measurements, such as the most recent or average value,
without unreliable higher-order temporal statistics. \\

\bottomrule
\end{tabular}
\end{table*}

\paragraph{Representative Laboratory Instruction.}

As a representative example, the following instruction instantiates the source-specific template for laboratory measurements in the Mortality-48h task. It is dynamically combined with the general DP system prompt during local execution.

\begin{promptbox}{Source-Specific Data Preprocessing Instruction: Laboratory}

\textbf{EHR SOURCE}

Laboratory

\textbf{DATA CHARACTERISTICS}

The laboratory source contains sparse and irregular longitudinal measurements with potentially high missingness. Different laboratory variables may have substantially different sampling frequencies, and
multiple observations of the same variable may occur within one temporal interval.

\textbf{AVAILABLE VARIABLES}

Representative variables include glucose, creatinine, bicarbonate, potassium, platelets, and other available laboratory measurements.

\textbf{SOURCE-SPECIFIC FEATURE CONSTRUCTION}

For each patient, temporal interval, and laboratory variable, construct a small set of informative and clinically interpretable features.

Candidate features may include:

\begin{promptitemize}
    \item a representative value, such as the most recent or mean measurement within the interval;
    \item temporal change relative to preceding observations when sufficient measurements are available;
    \item within-interval variability when repeated measurements are available.
\end{promptitemize}

Avoid excessive or redundant feature expansion. Do not construct temporal change or variability features when the available observations are insufficient to support reliable estimation.

\textbf{REPEATED AND MISSING OBSERVATIONS}

When multiple measurements of the same variable occur within one temporal interval, apply an appropriate aggregation strategy according to the characteristics of the variable.

When no measurement is available, use a stable strategy such as forward filling, an explicit missing representation, or an appropriate default value.

\textbf{FEATURE NAMING}

Use deterministic and interpretable feature names following:

\begin{center}
\texttt{\{variable\_name\}\_\{feature\_type\}\_t\{time\_index\}}
\end{center}

Examples include:

\begin{promptitemize}
    \item \texttt{glucose\_last\_t0}
    \item \texttt{glucose\_mean\_t0}
    \item \texttt{creatinine\_last\_t0}
    \item \texttt{potassium\_mean\_t0}
\end{promptitemize}

Ensure that the feature-name ordering exactly matches the flattened temporal representation.

\end{promptbox}

\paragraph{Data Preprocessing Evaluation.}

Following local execution, the generated preprocessing workflow is assessed by the DP Evaluator. The evaluator examines the executable workflow together with its structured execution evidence and local historical context, and produces textual feedback for subsequent local prompt refinement. It does
not directly modify the generated code.

\begin{promptbox}{System Prompt for Data Preprocessing Evaluation}

You are an expert in clinical EHR preprocessing acting as an evaluator for the current data preprocessing workflow.

Your task is to critically assess the generated preprocessing pipeline and provide actionable feedback that can improve subsequent preprocessing and downstream clinical prediction performance.

\textbf{DYNAMIC INPUTS}

You will be provided with:

\begin{promptitemize}
    \item \texttt{Clinical Task: \{TASK\_DESCRIPTION\}}
    \item \texttt{Target Outcome: \{TARGET\_OUTCOME\}}
    \item \texttt{Observation Window: \{OBSERVATION\_WINDOW\}}
    \item \texttt{Source Name: \{TABLE\_NAME\}}
    \item \texttt{Generated Preprocessing Code:
    \{PREPROCESSING\_CODE\}}
    \item \texttt{Feature Summary: \{FEATURE\_SUMMARY\}}
    \item \texttt{Data Quality Statistics:
    \{DATA\_QUALITY\_STATISTICS\}}
    \item \texttt{Execution Statistics:
    \{EXECUTION\_STATISTICS\}}
    \item \texttt{Validation Performance:
    \{VALIDATION\_PERFORMANCE\}}
    \item \texttt{Local Historical Context:
    \{LOCAL\_HISTORICAL\_CONTEXT\}}
\end{promptitemize}

The supplied information contains modeling-level summaries only. Do not request or infer patient-level EHR values.

Do not generate or rewrite executable code. Your role is evaluation and feedback only.

Organize the evaluation into four parts.

\textbf{PART 1: Historical Execution Analysis}

\begin{promptitemize}
    \item Compare the current preprocessing outcome with relevant previous local executions.
    \item Identify preprocessing decisions associated with previous improvements, degradation, stagnation, or abnormal outcomes.
    \item Determine whether the current workflow improves feature quality, data coverage, temporal representation, or downstream validation performance.
\end{promptitemize}

\textbf{PART 2: Preprocessing Correctness and Temporal Validity}

\begin{promptitemize}
    \item Verify that the preprocessing workflow is consistent with the specified clinical task and observation window.
    \item Detect potential target leakage, future-information leakage, or incorrect temporal alignment.
    \item Assess patient alignment and consistency of the temporal representation.
    \item Evaluate whether repeated, missing, and irregular observations are handled appropriately for the current EHR source.
\end{promptitemize}

\textbf{PART 3: Feature Quality, Clinical Relevance, and Robustness}

\begin{promptitemize}
    \item Assess whether the constructed features are clinically meaningful and relevant to the target prediction task.
    \item Evaluate whether temporal aggregation, feature construction, and feature selection are justified by the characteristics of the source.
    \item Identify redundant, unstable, excessively sparse, or weakly informative features.
    \item Examine whether missingness, measurement frequency, or source-specific recording patterns may introduce artifacts or reduce generalizability.
\end{promptitemize}

\textbf{PART 4: Refinement Recommendations}

\begin{promptitemize}
    \item Identify the highest-impact limitations of the current preprocessing strategy.
    \item Provide concrete and actionable recommendations for the next preprocessing iteration.
    \item Preserve strategies supported by positive execution or validation evidence.
    \item If performance has stagnated or degraded across previous iterations, identify likely causes and recommend meaningful changes rather than minor variations of unsuccessful strategies.
\end{promptitemize}

\textbf{OUTPUT}

Return a concise four-part evaluation containing actionable textual feedback for subsequent local prompt refinement. Do not output executable code.

\end{promptbox}

\subsection{Model Development Prompts}
\label{appendix:md_prompts}

The MD Module operates on the temporal feature representations generated by the DP Module and autonomously constructs the corresponding clinical prediction workflow. Because the prediction tasks
considered in this work are formulated over longitudinal EHR representations, the MD Module is instructed to prioritize models that explicitly capture temporal dependencies, while retaining conventional machine-learning models as alternatives when more appropriate for the local data.

The MD prompt consists of a shared system instruction together with task-, hospital-, and round-specific information dynamically inserted at runtime. In addition to the current local MD prompt and locally maintained historical context, the module receives the global MD meta-prompt generated
by the federated server in the preceding communication round. The global meta-prompt provides transferable cross-hospital modeling guidance, whereas the local prompt and historical context preserve hospital-specific modeling experience. For the initial communication round, the global meta-prompt is
empty.

\paragraph{General MD System Prompt.}
The following system prompt defines the common model-development requirements used across hospitals and clinical prediction tasks.

\begin{promptbox}{System Prompt for Model Development}

You are a clinical temporal modeling code generator.

Your task is to generate one complete and executable Python program for training, selecting, and evaluating a prediction model using processed longitudinal EHR representations.

\textbf{TASK OBJECTIVE}

Given the longitudinal EHR observations available within the specified observation window, develop a predictive model for the target clinical outcome defined by:

\begin{promptitemize}
    \item \texttt{Task: \{TASK\_NAME\}}
    \item \texttt{Target Outcome: \{TARGET\_OUTCOME\}}
    \item \texttt{Observation Window: \{OBSERVATION\_WINDOW\}}
    \item \texttt{Prediction Horizon: \{PREDICTION\_HORIZON\}}
\end{promptitemize}

The objective is to maximize predictive performance on the specified evaluation metric while maintaining robust generalization to unseen patients.

\textbf{DYNAMIC INPUTS}

You will be provided with:

\begin{promptitemize}
    \item \texttt{Training Data: \{TRAIN\_DATA\_PATH\}}
    \item \texttt{Validation Data: \{VALIDATION\_DATA\_PATH\}}
    \item \texttt{Test Data: \{TEST\_DATA\_PATH\}}
    \item \texttt{Input Shape: \{INPUT\_SHAPE\}}
    \item \texttt{Feature Metadata: \{FEATURE\_METADATA\}}
    \item \texttt{Class Statistics: \{CLASS\_STATISTICS\}}
    \item \texttt{Evaluation Metric: \{EVALUATION\_METRIC\}}
    \item \texttt{Global Meta-Prompt: \{GLOBAL\_MD\_META\_PROMPT\}}
\end{promptitemize}

The global meta-prompt contains transferable modeling guidance obtained from previous cross-hospital collaboration. Use it as guidance, but adapt the final modeling strategy according to the characteristics and validation evidence of the current local data.

\textbf{ROLE BOUNDARY}

\begin{promptitemize}
    \item Focus only on predictive model construction, training, hyperparameter optimization, evaluation, and interpretation.
    \item The input data have already been transformed into temporally structured model-ready representations.
    \item Do not redesign the underlying clinical feature-construction procedure.
    \item Apply only transformations required by the selected model, such as model-specific scaling, tensor conversion, or masking.
    \item Do not use patient identifiers as predictive features.
\end{promptitemize}

\textbf{MODEL SELECTION}

The input generally follows a temporal representation of shape $(N,T,D)$, where $N$ is the number of patients, $T$ is the number of temporal intervals, and $D$ is the number of feature channels.

Prioritize models that explicitly capture temporal dependencies, including:

\begin{promptitemize}
    \item Temporal Convolutional Networks (TCNs);
    \item Long Short-Term Memory networks (LSTMs);
    \item Gated Recurrent Units (GRUs) or other RNN variants;
    \item temporal Transformer-based models;
    \item other suitable lightweight temporal architectures.
\end{promptitemize}

Select the architecture according to the local sample size, sequence length, feature dimensionality, sparsity, class distribution, and validation performance.

If temporal models are unsuitable, unstable, or consistently inferior on the validation set, conventional models such as logistic regression, random forest, XGBoost, LightGBM, CatBoost, or multilayer perceptrons may be used as alternatives.

Do not prefer a model solely because it is more complex. Model selection must be supported by validation performance.

\textbf{TRAINING AND EVALUATION REQUIREMENTS}

\begin{promptenum}

    \item Load the training, validation, and test representations from the provided local paths.

    \item Preserve the temporal ordering of the input when using temporal models. Flatten the temporal dimension only when required by a non-temporal model.

    \item Use the training set for model fitting.

    \item Use the validation set for model selection, hyperparameter tuning, early stopping, and other optimization decisions.

    \item Do not use the test set for model selection or hyperparameter optimization.

    \item After determining the best model configuration, perform the final test evaluation only for reporting.

    \item Use \texttt{\{EVALUATION\_METRIC\}} as the primary model-selection criterion. For the imbalanced clinical prediction tasks considered in this work, AUPRC is used as the primary metric.

    \item Account for class imbalance using appropriate loss weighting, sampling strategies, or model-specific mechanisms when necessary.

    \item Apply reasonable hyperparameter exploration, regularization, and early stopping while avoiding unnecessary model complexity.

    \item Fix random seeds whenever possible to ensure reproducibility.

\end{promptenum}

\textbf{SHAP FEATURE IMPORTANCE}

SHAP-based feature attribution is required for the final selected model.

\begin{promptitemize}

    \item Compute SHAP values using the validation set after the best model has been selected.

    \item Compute the mean absolute SHAP value for each input feature.

    \item Rank features in descending order of mean absolute SHAP value.

    \item Preserve the correspondence between SHAP values and the provided
    feature names.

    \item Save the complete feature-importance results to \texttt{\{SHAP\_OUTPUT\_PATH\}}.

    \item Generate a SHAP summary or beeswarm plot for the most influential features and save it to
    \texttt{\{SHAP\_FIGURE\_PATH\}}.

    \item SHAP results must not influence model selection or test-set evaluation.

    \item If the primary SHAP explainer is incompatible with the selected model, use an appropriate alternative SHAP explainer. If SHAP computation ultimately fails, explicitly record the failure rather than silently omitting the output.

\end{promptitemize}

\textbf{REQUIRED OUTPUTS}

The generated program must save structured modeling outputs containing:

\begin{promptitemize}
    \item selected model family and configuration;
    \item training performance;
    \item validation performance;
    \item final test performance;
    \item selected hyperparameters;
    \item relevant training and convergence statistics;
    \item complete SHAP feature-importance results;
    \item execution status and relevant runtime statistics.
\end{promptitemize}

All outputs must contain modeling-level information only and must not expose patient-level EHR records.

\textbf{OUTPUT CONSTRAINT}

Return only one executable Python program. Do not include explanations, Markdown text, or any content outside the generated code.

\end{promptbox}

\paragraph{Model Development Evaluation.}

Following local model execution, the MD Evaluator assesses the generated modeling workflow using its training behavior, validation performance, feature-attribution results, and local historical context. The evaluator does not directly modify the generated code; instead, it produces structured feedback for subsequent local prompt refinement.

\begin{promptbox}{System Prompt for Model Development Evaluation}

You are an expert in clinical temporal modeling acting as an evaluator for the current model-development workflow.

Your task is to critically assess the generated modeling pipeline and provide actionable feedback that can improve predictive performance, temporal modeling quality, and generalization on the specified clinical task.

\textbf{DYNAMIC INPUTS}

You will be provided with:

\begin{promptitemize}

    \item \texttt{Clinical Task: \{TASK\_DESCRIPTION\}}

    \item \texttt{Target Outcome: \{TARGET\_OUTCOME\}}

    \item \texttt{Observation Window: \{OBSERVATION\_WINDOW\}}

    \item \texttt{Prediction Horizon: \{PREDICTION\_HORIZON\}}

    \item \texttt{Input Shape: \{INPUT\_SHAPE\}}

    \item \texttt{Generated Modeling Code:
    \{MODELING\_CODE\}}

    \item \texttt{Selected Model:
    \{MODEL\_CONFIGURATION\}}

    \item \texttt{Training Performance:
    \{TRAIN\_PERFORMANCE\}}

    \item \texttt{Validation Performance:
    \{VALIDATION\_PERFORMANCE\}}

    \item \texttt{Training Statistics:
    \{TRAINING\_STATISTICS\}}

    \item \texttt{SHAP Feature Importance:
    \{SHAP\_SUMMARY\}}

    \item \texttt{Execution Statistics:
    \{EXECUTION\_STATISTICS\}}

    \item \texttt{Local Historical Context:
    \{LOCAL\_HISTORICAL\_CONTEXT\}}

\end{promptitemize}

Do not rewrite the generated modeling code. Your role is evaluation and feedback only.

Organize the evaluation into four parts.

\textbf{PART 1: Historical Modeling Analysis}

\begin{promptitemize}

    \item Compare the current modeling outcome with relevant previous local executions.

    \item Identify model families, architectures, or hyperparameter settings associated with previous improvements, degradation, stagnation, or unstable behavior.

    \item Detect recurring failure modes such as severe overfitting, underfitting, convergence failure, unstable validation performance, or ineffective class-imbalance handling.

    \item Determine whether the current workflow represents a meaningful improvement over previous local modeling strategies.

\end{promptitemize}

\textbf{PART 2: Modeling Correctness and Temporal Validity}

\begin{promptitemize}

    \item Verify that the modeling workflow is consistent with the clinical task, observation window, prediction horizon, and evaluation protocol.

    \item Assess whether the selected model appropriately exploits the temporal structure of the input representation.

    \item Check whether temporal ordering is preserved for sequence-based models and identify inappropriate flattening or loss of temporal information.

    \item Evaluate whether the selected temporal architecture is suitable for the available sample size, sequence length, feature dimensionality, sparsity, and class distribution.

    \item Detect potential leakage from test-set usage, future observations, or inappropriate model-selection procedures.

\end{promptitemize}

\textbf{PART 3: Model Selection, Training Behavior, and Feature Attribution}

\begin{promptitemize}

    \item Assess whether the selected model family and hyperparameters are justified by validation performance.

    \item Compare the selected temporal model with plausible alternative architectures and determine whether its complexity is warranted.

    \item Examine training and validation behavior for evidence of overfitting, underfitting, poor convergence, or excessive model complexity.

    \item Evaluate whether class imbalance, regularization, early stopping, and hyperparameter exploration are handled appropriately.

    \item Examine the SHAP feature-importance results and determine whether the influential features are clinically plausible and consistent with the prediction task.

    \item Identify suspicious attribution patterns that may indicate leakage, spurious correlations, unstable features, or poor utilization of the temporal representation.

\end{promptitemize}

\textbf{PART 4: Refinement Recommendations}

\begin{promptitemize}

    \item Identify the highest-impact limitations of the current modeling strategy.

    \item Provide concrete recommendations for model architecture, hyperparameters, optimization, regularization, or temporal modeling.

    \item Preserve modeling strategies that are consistently supported by positive validation evidence.

    \item If validation performance has stagnated or degraded across previous iterations, analyze the likely causes and recommend meaningful changes to the model family or training strategy.

    \item Prioritize improvements that are likely to generalize rather than increasing model complexity without supporting evidence.

\end{promptitemize}

\textbf{OUTPUT}

Return a concise four-part evaluation containing actionable textual feedback for subsequent local prompt refinement. Do not output executable code.

\end{promptbox}

\subsection{Shared Local Execution and Refinement}
\label{appendix:shared_prompts}

The DP and MD modules share the same local execution, code-repair, and prompt-refinement infrastructure. Generated programs are first executed within the local hospital environment. The Code Executor is a deterministic runtime component rather than an LLM-based component and therefore does not
require a separate prompt. It records the execution status, standard output, runtime diagnostics, and exception traces. When execution fails, these artifacts are provided to the Code Repairer for targeted correction.

After successful execution and module-specific evaluation, the resulting textual feedback is used by TextGrad to refine the corresponding local DP or MD prompt. This refinement is performed locally and does not require an additional task-specific prompt template beyond the TextGrad optimization
procedure. Historical memory used during refinement also remains within the originating hospital.

\paragraph{Code Repair.}

The Code Repairer is shared by the DP and MD modules. Its objective is to resolve execution failures while preserving the intended preprocessing or modeling logic of the generated program.

\begin{promptbox}{System Prompt for Code Repair}

You are a Python code repairer for automated clinical EHR modeling.

Your task is to identify the cause of an execution failure and return a corrected version of the generated program.

\textbf{DYNAMIC INPUTS}

You will be provided with:

\begin{promptitemize}
    \item \texttt{Task Context: \{TASK\_CONTEXT\}}
    \item \texttt{Module Type: \{MODULE\_TYPE\}}
    \item \texttt{Generated Code: \{GENERATED\_CODE\}}
    \item \texttt{Execution Error: \{ERROR\_TRACE\}}
    \item \texttt{Runtime Diagnostics: \{RUNTIME\_DIAGNOSTICS\}}
    \item \texttt{Required Outputs: \{OUTPUT\_REQUIREMENTS\}}
\end{promptitemize}

\textbf{REPAIR REQUIREMENTS}

\begin{promptitemize}

    \item Identify the direct cause of the execution failure.

    \item Apply the minimum necessary correction required to restore successful execution.

    \item Preserve the original preprocessing or modeling objective whenever possible.

    \item Do not introduce unrelated feature-engineering, model-selection, or methodological changes.

    \item Preserve required input paths, output paths, data interfaces, tensor dimensions, and feature ordering.

    \item Preserve temporal-validity, leakage-prevention, and patient-alignment constraints defined by the original task.

    \item Avoid unnecessary dependencies or substantial restructuring of the generated program.

\end{promptitemize}

If multiple execution errors are present, resolve them systematically while preserving the original workflow semantics.

\textbf{OUTPUT CONSTRAINT}

Return only one corrected executable Python program. Do not include explanations, Markdown text, or additional content outside the code.

\end{promptbox}

\paragraph{TextGrad-Based Local Prompt Refinement.}

Local prompt refinement is performed using TextGrad rather than a separate LLM prompt template. For each module $m\in\{\mathrm{DP},\mathrm{MD}\}$, the evaluator produces textual feedback that identifies weaknesses in the current execution and provides actionable recommendations. TextGrad treats this feedback as a textual optimization signal and propagates it to the corresponding executable prompt.

Specifically, given the current local prompt $\mathcal{P}_{k,m}^{(r)}$, evaluator feedback $\mathcal{F}_{k,m}^{(r)}$, and relevant locally maintained historical context, TextGrad updates the prompt to obtain the locally refined prompt
$
\widetilde{\mathcal{P}}_{k,m}^{(r)}
=
\operatorname{TextGrad}
\left(
\mathcal{P}_{k,m}^{(r)},
\mathcal{F}_{k,m}^{(r)},
\mathcal{M}_{k}^{(r)}
\right).
$

The refinement process aims to preserve strategies supported by positive execution or validation evidence while correcting weaknesses identified by the evaluator. Historical observations are used to discourage repeated unsuccessful decisions and to retain previously effective strategies. Importantly, this optimization modifies the executable DP or MD prompt rather than directly rewriting the generated program.

All TextGrad refinement and historical memory remain local to each hospital. Only the subsequently constructed module-specific experience representation, consisting of the refined prompt and structured modeling evidence, is communicated to the federated server.

\subsection{Federated Experience Aggregation Prompts}
\label{appendix:federated_prompts}

After local execution, evaluation, and TextGrad-based prompt refinement, each hospital constructs separate experience representations for the DP and MD modules. Each representation contains the locally refined executable prompt together with structured modeling evidence derived from local execution. Historical memory, generated code, and patient-level EHR records remain local and are not included in the uploaded representation.

At the federated server, Evidence-Guided Experience Aggregation (EGEA) first integrates the uploaded module-specific experiences according to their supporting evidence. The resulting aggregated experience is then converted into a concise global meta-prompt, which is returned to participating hospitals as transferable guidance for the next communication round.

\paragraph{Experience Aggregation.}
The following prompt guides the server-side integration of module-specific local experiences using their associated structured modeling evidence.

\begin{promptbox}{System Prompt for Experience Aggregation}

You are responsible for aggregating clinical modeling experience collected from multiple hospitals for a specified functional module.

Your objective is to identify reliable and transferable modeling knowledge from heterogeneous local experiences while preserving useful context-dependent strategies.

\textbf{DYNAMIC INPUTS}

You will be provided with:

\begin{promptitemize}

    \item \texttt{Clinical Task: \{TASK\_DESCRIPTION\}}

    \item \texttt{Module Type: \{MODULE\_TYPE\}}

    \item \texttt{Communication Round: \{ROUND\}}

    \item \texttt{Local Experience Representations:
    \{LOCAL\_EXPERIENCE\_REPRESENTATIONS\}}

\end{promptitemize}

Each local experience representation contains:

\begin{promptitemize}
    \item a locally refined executable prompt;
    \item structured modeling evidence obtained from local execution and evaluation.
\end{promptitemize}

Historical memory, executable code, and patient-level EHR records are not provided.

\textbf{AGGREGATION REQUIREMENTS}

\begin{promptitemize}

    \item Assess each local modeling strategy according to the evidence supporting its effectiveness, stability, and task relevance.

    \item Identify strategies that are consistently supported by reliable evidence across multiple hospitals.

    \item Preserve complementary strategies when they provide useful information under different local data characteristics.

    \item Distinguish broadly transferable strategies from strategies whose effectiveness depends on specific data properties, such as sample size, feature availability, temporal sparsity, or class imbalance.

    \item Identify conflicting local strategies and determine whether the differences can be explained by heterogeneous data characteristics.

    \item Downweight or exclude strategies associated with weak, deteriorating, unstable, or contradictory evidence.

    \item Do not use simple prompt averaging, majority voting, or direct concatenation as the aggregation criterion.

    \item Do not infer, reconstruct, or expose patient-level information.

\end{promptitemize}

When \texttt{\{MODULE\_TYPE\}} corresponds to data preprocessing, focus on transferable experience related to temporal construction, source-aware feature engineering, missingness handling, and leakage-safe preprocessing. 

When \texttt{\{MODULE\_TYPE\}} corresponds to model development, focus on transferable experience related to temporal model selection, optimization, generalization, class-imbalance handling, and feature attribution.

\textbf{OUTPUT STRUCTURE}

Return an aggregated experience representation containing:

\begin{promptitemize}

    \item \textbf{Supported Strategies:} modeling strategies consistently supported by reliable evidence;

    \item \textbf{Complementary Strategies:} additional transferable strategies supported under heterogeneous settings;

    \item \textbf{Context-Dependent Strategies:} strategies that should be applied only under particular data characteristics;

    \item \textbf{Strategies to Avoid:} weak, unstable, or conflicting strategies not supported by sufficient evidence;

    \item \textbf{Evidence Summary:} concise justification for the aggregation decisions.

\end{promptitemize}

Return only the aggregated experience representation.

\end{promptbox}

\paragraph{Global Meta-Prompt Generation.}

The aggregated experience is subsequently transformed into a module-specific global meta-prompt. Unlike local TextGrad refinement, this step does not modify a hospital's local prompt directly. Instead, it distills cross-hospital experience into transferable guidance that is dynamically provided to the corresponding DP or MD Generator in the next communication round.

\begin{promptbox}{System Prompt for Global Meta-Prompt Generation}

You are responsible for converting aggregated federated clinical modeling experience into a concise and transferable global meta-prompt.

\textbf{DYNAMIC INPUTS}

You will be provided with:

\begin{promptitemize}

    \item \texttt{Clinical Task: \{TASK\_DESCRIPTION\}}

    \item \texttt{Module Type: \{MODULE\_TYPE\}}

    \item \texttt{Communication Round: \{ROUND\}}

    \item \texttt{Aggregated Experience:
    \{AGGREGATED\_EXPERIENCE\}}

\end{promptitemize}

\textbf{OBJECTIVE}

Generate actionable cross-hospital guidance that can be incorporated into the corresponding local DP or MD modeling process in the next communication round.

The global meta-prompt should communicate transferable experience rather than prescribe one fixed preprocessing pipeline, feature set, model architecture, or hyperparameter configuration.

\textbf{GENERATION REQUIREMENTS}

\begin{promptitemize}

    \item Prioritize strategies supported by strong and consistent cross-hospital evidence.

    \item Preserve useful complementary strategies when their effectiveness depends on local data characteristics.

    \item Express context-dependent recommendations together with the conditions under which they are likely to be useful.

    \item Exclude strategies associated with weak, unstable, or contradictory evidence.

    \item Provide actionable guidance while allowing each hospital to adapt the final implementation according to its local data and validation evidence.

    \item Avoid enforcing an identical feature set, preprocessing workflow, temporal model, or hyperparameter configuration across hospitals.

    \item Do not include patient-level information, generated code, historical memory, or hospital identifiers.

    \item Keep the resulting guidance concise and directly usable as a dynamic input to the corresponding local module.

\end{promptitemize}

When generating a data pre-processing meta-prompt, emphasize transferable preprocessing strategies, including temporal representation, source-aware feature construction, missingness handling, and leakage prevention.

When generating a model development meta-prompt, emphasize transferable modeling strategies, including temporal model selection, optimization, generalization, class-imbalance handling, and interpretation.

\textbf{OUTPUT STRUCTURE}

Organize the global meta-prompt into:

\begin{promptitemize}

    \item \textbf{Recommended Strategies:} broadly supported guidance for the next communication round;

    \item \textbf{Conditional Guidance:} strategies that should be considered only under specified local data characteristics;

    \item \textbf{Warnings:} strategies or failure modes that should be avoided.

\end{promptitemize}

\textbf{OUTPUT CONSTRAINT}

Return only the global meta-prompt. Do not include analysis, explanations, patient-level information, or executable code.

\end{promptbox}

\subsection{Representative Structured Outputs}
\label{appendix:structured_outputs}

FedEHR-Agents organizes local execution outcomes into compact structured modeling evidence for subsequent evaluation and federated experience aggregation. The following representative schemas illustrate the module-specific information included in the shared experience representations and the structured outputs generated at the federated server. The placeholders specify the information interface and do not represent patient-level observations or local historical memory.

\begin{promptbox}{Structured Output Schema for DP Modeling Evidence}

\textbf{Feature Construction}

\texttt{selected\_features:}
\texttt{[<feature\_1>, <feature\_2>, ..., <feature\_n>]}

\texttt{num\_selected\_features: <integer>}

\texttt{feature\_source\_summary: <grouped summary>}

\textbf{Temporal and Data Quality}

\texttt{temporal\_coverage: <summary>}

\texttt{missingness\_summary: <summary>}

\texttt{temporal\_consistency: <assessment>}

\textbf{Downstream Validation Evidence}

\texttt{validation\_metric: <metric name>}

\texttt{validation\_score: <value>}

\textbf{Execution Evidence}

\texttt{execution\_status: <status>}

\texttt{execution\_time: <value>}

\end{promptbox}

\begin{promptbox}{Structured Output Schema for MD Modeling Evidence}

\textbf{Model Configuration}

\texttt{model\_family: <selected model>}

\texttt{model\_architecture: <architecture summary>}

\texttt{hyperparameters: <model-specific configuration>}

\textbf{Predictive Performance}

\texttt{training\_performance: <value>}

\texttt{validation\_performance: <value>}

\texttt{evaluation\_metric: <metric name>}

\textbf{Training Behavior}

\texttt{convergence\_status: <assessment>}

\texttt{overfitting\_assessment: <assessment>}

\texttt{class\_imbalance\_handling: <strategy>}

\textbf{SHAP Feature Attribution}

\texttt{shap\_feature\_ranking:}
\texttt{[(<feature>, <importance>), ...]}

\texttt{top\_shap\_features:}
\texttt{[<feature\_1>, ..., <feature\_k>]}

\textbf{Execution Evidence}

\texttt{execution\_status: <status>}

\texttt{execution\_time: <value>}

\end{promptbox}

\begin{promptbox}{Structured Output Schema for Aggregated Experience}

\textbf{Supported Strategies}

\texttt{<strategies consistently supported by reliable cross-hospital evidence>}

\textbf{Complementary Strategies}

\texttt{<transferable strategies supported under heterogeneous settings>}

\textbf{Context-Dependent Strategies}

\texttt{<strategies applicable under specific local data characteristics>}

\textbf{Strategies to Avoid}

\texttt{<weak, unstable, or contradictory strategies>}

\textbf{Evidence Summary}

\texttt{<concise justification based on structured local modeling evidence>}

\end{promptbox}

\begin{promptbox}{Structured Output Schema for Global Meta-Prompt}

\textbf{Recommended Strategies}

\texttt{<transferable guidance supported by cross-hospital evidence>}

\textbf{Conditional Guidance}

\texttt{<strategies applicable under specified local data characteristics>}

\textbf{Warnings}

\texttt{<unsupported strategies or failure modes to avoid>}

\end{promptbox}

\end{document}